\documentclass[10pt,twocolumn,letterpaper]{article}

\usepackage[pagenumbers]{cvpr} 

\definecolor{cvprblue}{rgb}{0.21,0.49,0.74}
\usepackage[pagebackref,breaklinks,colorlinks,allcolors=cvprblue]{hyperref}

\def\paperID{1997} 
\def\confName{CVPR}
\def\confYear{2026}

\title{ParticleGS: Learning Neural Gaussian Particle Dynamics from Videos \\
for Prior-free Physical Motion Extrapolation}

\author{
Jinsheng Quan$^{1}$, Qiaowei Miao$^{1}$, Yichao Xu$^{1}$, Zizhuo Lin$^{1}$, Ying Li$^{2}$, Wei Yang$^{3}$, Zhihui Li$^{4}$, Yawei Luo$^{1}$\textsuperscript{~\Letter} \\
$^{1}$Zhejiang University \quad $^{2}$North China University of Technology\\
$^{3}$Huazhong University of Science and Technology \quad $^{4}$University of Science and Technology of China\\
\vspace{-0.6cm}
}

\begin{document}
\vspace{-0.5cm}
\twocolumn[{
\maketitle
\begin{center}
    \includegraphics[width=\linewidth]{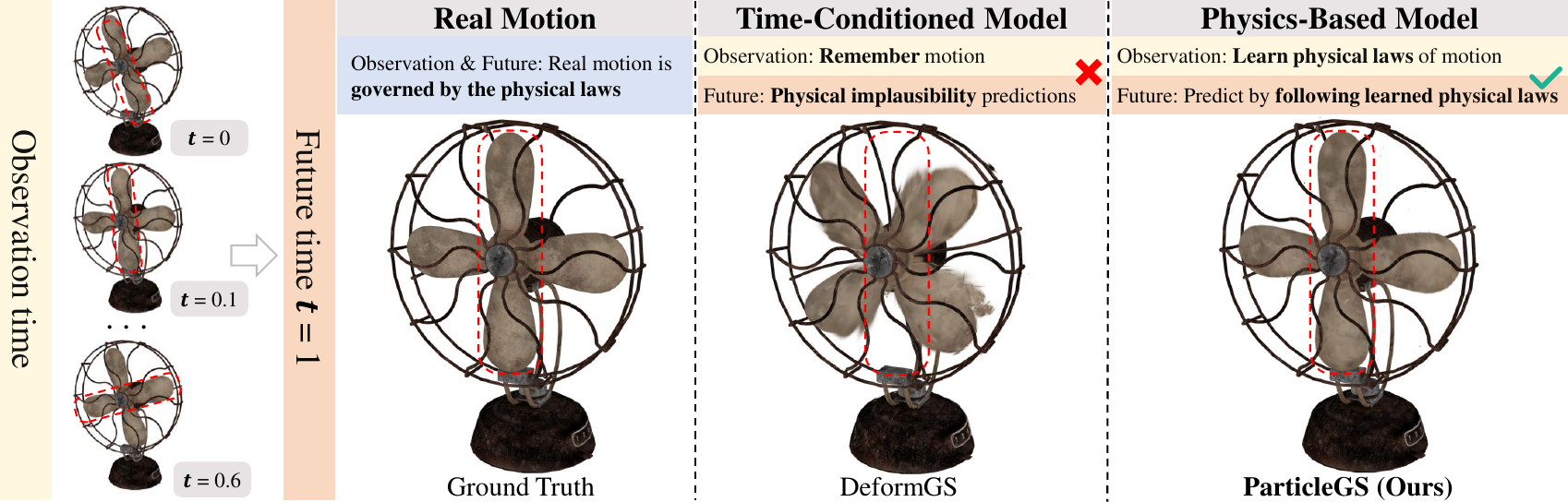}
    \end{center}
    \vspace{-0.4cm}
    \captionsetup{type=figure}
    \captionof{figure}{\textbf{Example of typical reconstruction method and our method for extrapolation}. Given multi-view RGB video observations, our \textbf{ParticleGS} can learn latent physical dynamics and use them to extrapolate future Gaussians with physically consistent motion and appearance. In contrast, existing time-conditioned deformation methods fail to predict plausible future motion.}
    \label{firstfig}
    \vspace{0.2cm}
}]

\begin{abstract}
The ability to extrapolate dynamic 3D scenes beyond the observed timeframe is fundamental to advancing physical world understanding and predictive modeling. Existing dynamic 3D reconstruction methods have achieved high-fidelity rendering of temporal interpolation, but typically lack physical consistency in predicting the future. To overcome this issue, we propose ParticleGS, a physics-based framework that reformulates dynamic 3D scenes as physically grounded systems. ParticleGS comprises three key components: 1) an encoder that decomposes the scene into static properties and initial dynamic physical fields; 2) an evolver based on Neural Ordinary Differential Equations (Neural ODEs) that learns continuous-time dynamics for motion extrapolation; and 3) a decoder that reconstructs 3D Gaussians from evolved particle states for rendering. Through this design, ParticleGS integrates physical reasoning into dynamic 3D representations, enabling accurate and consistent prediction of the future. Experiments show that ParticleGS achieves state-of-the-art performance in extrapolation while maintaining rendering quality comparable to leading dynamic 3D reconstruction methods. 
\vspace{-1\baselineskip}
\end{abstract}
 
\section{Introduction}
\label{sec:intro}
Learning the physical dynamics from multi-view videos and inferring the subsequent motion is crucial for applications like gaming, autonomous driving, and robotics~\cite{wang2024driving,yin2021modeling}. Recent methods~\cite{deformable3d,4dgs,dnerf,hexplane,grid4d,TiVox,miao2025advances} have demonstrated impressive results in reconstructing and rendering vivid dynamic 3D scenes. Nevertheless, these methods typically predict deformation as a function of time. While this formulation suffices for interpolating motion within the observed temporal range, it fails to learn the underlying physical laws. This results in unsatisfactory performance when extrapolating to unseen future states, as shown in Fig. \ref{firstfig}.

To overcome these limitations, incorporating physical laws into dynamic 3D scene modeling has emerged as a promising direction~\cite{neurofluid,cao2024neuma,daviet2024neurally,sharma2022accelerated}. Existing approaches can be broadly categorized into two types. The first explicitly injects \textbf{physics priors} through simulation frameworks~\cite{physgaussian,feng2024pie,guo2024physically, omniphysgs}, physics-informed neural networks~\cite{chu2022physics,raissi2019physics,nvfi,cuomo2022scientific}, or architectures with built-in physical constraints~\cite{gsp,nvfi}. While enhancing physical plausibility, such methods often depend on manually specified external forces or restrictive assumptions, limiting their flexibility and generalization. The second line of work inherently relies on \textbf{geometric priors}, such as pre-processed point clouds or meshes, to infer physical information~\cite{niemeyer2019occupancy,jiang2021learning,ma2023learning,zhongsymplectic}. However, these methods cannot directly capture dynamics from raw observations and typically require multi-stage optimization. More recent works have modeled Gaussian rigid motion using velocity and acceleration fields~\cite{li2025freegave,li2025trace}. Nevertheless, such low-order dynamics are insufficient to accurately capture complex deformations, making it still challenging to learn physical dynamics directly from raw observations. 

In this paper, we propose ParticleGS, a dynamic 3D representation framework based on 3DGS~\cite{3dgs} from the perspective of physical particle dynamics. Our key insight is to treat each Gaussian as a particle, whose temporal evolution is driven by \textit{\textbf{physical state vector}} governed by the underlying physical laws. To this end, ParticleGS comprises three key components: 1) A Dynamics Latent Space Encoder that maps each Gaussian to an initial latent physical state vector, which is factorized into static properties (\textit{e.g.}, mass, material) and initial time-varying dynamic properties (\textit{e.g.}, velocity, acceleration). 2) A Neural ODE-based Dynamics Evolver, the core of our approach, which learns a continuous Markov state transition function in the latent space to approximate the underlying physical dynamics (\textit{e.g.}, Newtonian or Navier–Stokes) governing the particles, rather than merely memorizing temporal deformations. 3) A Gaussian Space Decoder that translates the evolved latent states back into the deformation of each Gaussian for rendering. By focusing on learning the underlying dynamics rather than memorizing temporal deformation, ParticleGS achieves extrapolation beyond observations. 

Our neural dynamics modeling distinguishes ParticleGS from existing approaches. Compared with time-conditioned reconstruction methods~\cite{4dgs,grid4d,deformable3d}, ParticleGS learns neural dynamics directly from videos, enabling physically consistent extrapolation beyond observed frames. Unlike methods that rely on physical or geometric priors~\cite{nvfi,physgaussian}, it requires no predefined physical laws or structured geometric inputs, achieving prior-free dynamic learning. In contrast to approaches that model specific low-order physical quantities, such as velocity fields~\cite{li2025freegave,li2025trace}, ParticleGS learns a generalized formulation of higher-order particle dynamics through neural ordinary differential equations. The main contributions are summarized as follows:
\begin{itemize}[leftmargin=*, nosep, noitemsep]
    \item From a physical perspective, we reformulate time-conditioned deformation using neural ordinary differential equations, enabling a physics-grounded representation of the driving dynamics.
    \item We propose ParticleGS, an Encoder–Evolver–Decoder framework that represents the scene as a physical particle system with static properties and initial dynamic fields, where a Neural ODE-based evolver learns and extrapolates the latent physical laws directly from observations.
    \item Numerical experiments show that our method achieves the best performance in future Gaussian extrapolation on multiple datasets.
\end{itemize}

\section{Related Work}
\label{sec:related}
\textbf{Time-Conditioned Dynamic 3D Reconstruction.} Novel view synthesis for dynamic scenes has been significantly advanced by Neural Radiance Fields (NeRF)~\cite{nerf} and 3D Gaussian Splatting (3DGS)~\cite{3dgs}. To model temporal variations, a common strategy is to learn a deformation field that maps a canonical representation to the scene's state at any given time \cite{dnerf,park2021nerfies,deformable3d}. For NeRF-based methods, significant research has focused on improving representation efficiency and quality. This includes decomposing the 4D spacetime into lower-dimensional components like planes~\cite{fridovich2023k, hexplane, shao2023tensor4d} and employing compact yet expressive features such as hash encodings~\cite{muller2022instant, TiVox}. Following a similar trajectory, dynamic 3DGS methods~\cite{luiten2024dynamic, wuswift4d, kratimenos2024dynmf, huang2024sc, sun20243dgstream,bae2024per,lin2024gaussian,ma2024reconstructing,katsumata2024compact} extend static Gaussian clouds with learned motion, enabling high-fidelity, real-time rendering. 

\textbf{Injecting Physical Priors into Dynamic 3D.} To endow 3D representations with predictive power, integrating physical principles is a promising direction. One line of work follows the Physics-Informed Neural Network (PINN) paradigm~\cite{raissi2019physics}, embedding governing equations like the Navier-Stokes equations directly into the learning objective to constrain motion~\cite{nvfi,chu2022physics,im2024regularizing,haobi}. While effective, this approach is often confined to systems with well-defined physical laws and may not generalize to complex scenes. Another strategy involves integrating differentiable physics simulators~\cite{li2024diffavatar,du2021diffpd,liang2020differentiable, de2018end}, which provide strong physical priors. In addition, some methods embed rigid-body constraints, springs, or incorporate Graph Neural Networks~\cite{liu2024physgen} to embed neighborhood constraints~\cite{sanchez2020learning, zhong2024reconstruction, cao2023efficient, gsp}. 

\textbf{Learning Physical Information from Vision.} Inferring physical laws and properties directly from visual observations has attracted increasing interest. One line of work focuses on recovering object-level physical attributes, such as material properties~\cite{pacnerf,gic,feng2024pie,omniphysgs}, physically plausible geometry~\cite{guo2024physically,xiong2025topogaussian}, and velocity fields for motion extrapolation~\cite{li2025freegave,li2025trace}. Another line aims to learn underlying physical equations, including neural constitutive models~\cite{ma2023learning} and neural dynamics equations~\cite{lin2025visionlaw,neurofluid,cao2024neuma}. However, existing Neural ODE-based methods \cite{niemeyer2019occupancy,jiang2021learning,ma2023learning} struggle to learn particle-level dynamics purely from RGB videos. 

\textbf{Neural Ordinary Differential Equations.}  Unlike traditional discrete-layer architectures, Neural ODEs~\cite{ode} integrate continuous-time dynamical systems with neural networks. Specifically, they use a neural network $f$ to learn the derivative of the system's state. By employing an adaptive ODE solver to integrate this learned function over time, these models enable the learning of state evolution. This makes them well-suited for modeling physical processes.

\begin{figure*}
    \centering
    \includegraphics[width=\linewidth]{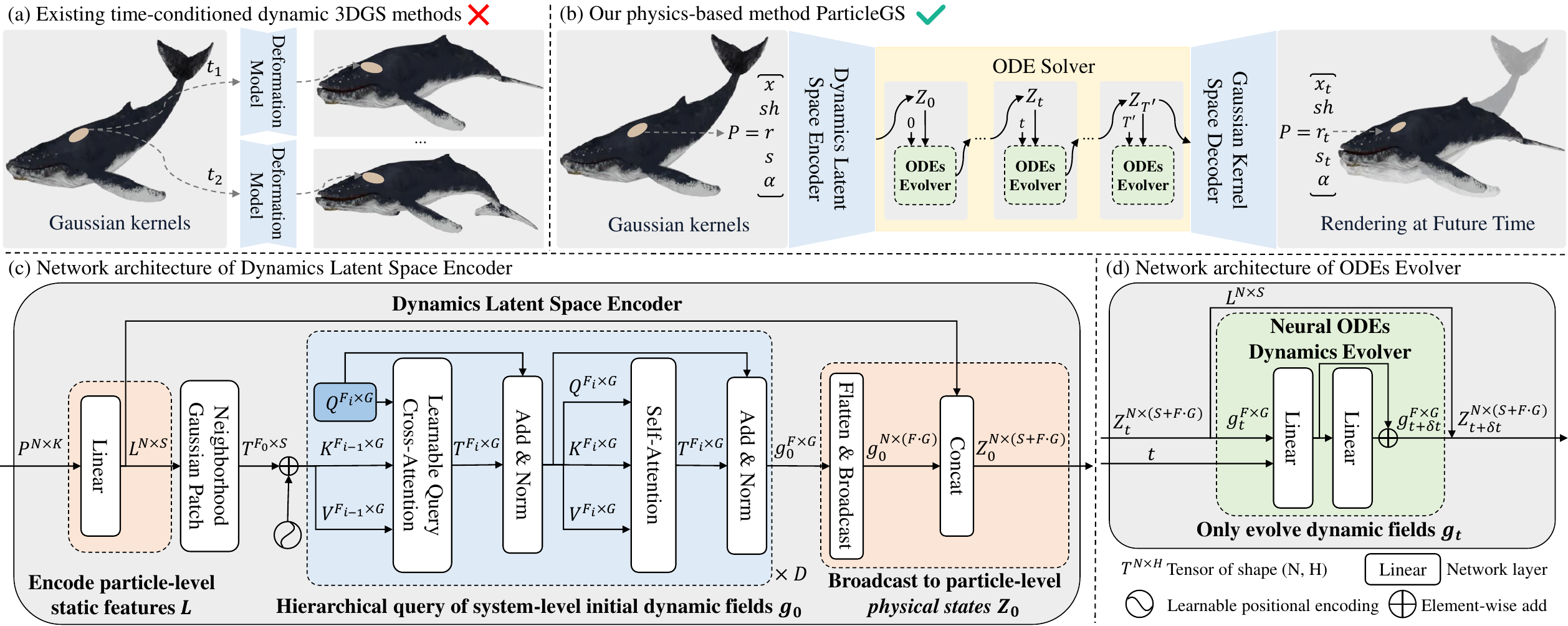}
    \caption{\textbf{Overview of ParticleGS}. (a) Existing time-conditioned methods learn a deformation model for each independent discrete time. (b) Our physics-based framework, ParticleGS, uses latent physical states to drive 3D Gaussian deformation, enabling motion extrapolation. (c) The Dynamics Latent Space Encoder maps Gaussians to an initial physical state, factorized into static properties and dynamic fields. (d) The Neural ODEs Evolver learns the underlying physical laws by modeling the continuous-time evolution of the dynamic fields.}
    \vspace{-0.5cm}
    \label{overview}
\end{figure*}
\section{Methodology} 
We begin by formulating the dynamic scene representation and positioning our physics-based approach against existing methods in Sec.~\ref{formulation}. We then detail the architecture of our framework, ParticleGS, in Sec.~\ref{particlegs} and describe its optimization strategy in Sec.~\ref{optimize}.
\subsection{Problem Formulation and Overview}
\label{formulation}
\textbf{3D Gaussian Splatting for Static Scenes.}
3D Gaussian Splatting~\cite{3dgs} aims to learn a set of Gaussian kernels $P=\{p_i\}_{i=1}^{N}$ from a set of images taken from corresponding camera viewpoints. Given a differentiable renderer $\mathcal{R}$ and a viewpoint $v$, an image $\hat{I}^v$ is rendered as:
\begin{equation}
    \mathcal{R}(P,v) = \hat{I}^v .
\end{equation}
Here, each Gaussian kernel $p$ is defined by a set of optimizable parameters $\{x, sh, r, s, \alpha\}$, with $x$ as position, $sh$ as Spherical Harmonics coefficients, $r$ as rotation, $s$ as scaling, and $\alpha$ as opacity. The Gaussian set is typically initialized from an SfM reconstruction~\cite{schonberger2016structure} or a random point cloud.

\textbf{Time-Conditioned Dynamic Scene Representation.} 
To model dynamic scenes, existing methods extend the static 3DGS framework. For a set of videos captured from corresponding viewpoints over $T$ discrete time steps, the primary goal is to learn a representation that can render the scene. These methods~\cite{deformable3d,grid4d,4dgs} typically learn a canonical set of 3D Gaussians $P$ and a time-conditioned deformation model $\mathcal{D}(P, t)=P_t$, which deforms $P$ into the Gaussian set $P_t$ at time $t$, as shown in Fig.~\ref{overview}(a). The image at time $t \in [0, T]$ and a viewpoint $v$ is rendered by:
\begin{equation}
    \mathcal{R}(\mathcal{D}(P,t),v) = \hat{I}_t^v .
    \label{t2D}
\end{equation}

However, these methods lack the modeling of underlying physical laws, which limits their ability to extrapolate. To address this, we propose to model the dynamic 3D scene as a system of particles governed by neural physical dynamics.

\textbf{Physical Dynamics-Based Representation (Ours).} For the same input, we employ the physical state set $Z_t$ to drive the Gaussian particles at time $t$. Under this definition, an image at time $t \in [0, T']$ (where $T'$ may extend beyond the observed time $T$) and a viewpoint $v$ is rendered as:
\begin{equation}
        \mathcal{R}(D(P,Z_t),v)=\hat{I}_t^v . 
        \label{z2D}
\end{equation}
Here, $D$ is a decoder that deforms the canonical Gaussians. The state set $Z_t$ is evolved from an initial state $Z_0$, which is encoded from the canonical Gaussians $P$.

Comparing Eq.~\ref{z2D} with Eq.~\ref{t2D}, our method drives the 3D Gaussians by the physical state rather than merely the timestamp $t$. This brings advantages: 1) Spatiotemporal awareness: $Z_t$ implicitly encodes the entire system history from $0$ to $t$, while $t$ encodes temporal order only; 2) Physical plausibility: Conditioning on a state $Z_t$ is more consistent with physical particle systems; 3) Markov property: The evolution from $Z_t$ is Markovian, as $Z_t$ contains all necessary information to predict the next state. Time-conditioned models lack this, as timestamps $t_1$ and $t_2$ are independent.

The framework of ParticleGS is shown in Fig.~\ref{overview}(b). ParticleGS consists of three components: 1) a Dynamics Latent Space Encoder that maps Gaussians $P$ to their initial physical states $Z_0$; 2) a Neural ODE-based Dynamics Evolver that models continuous dynamics to evolve $Z_0$ to $Z_t$ using a Neural ODE solver; and 3) a Gaussian Kernel Space Decoder that maps the evolved physical states $Z_t$ to the deformation of canonical Gaussian $P$ to form $P_t$ for rendering.

\subsection{ParticleGS}
\label{particlegs}
\textbf{Dynamics Latent Space Encoder.} 
Unlike existing methods that encode each Gaussian independently, we propose a Dynamics Latent Space Encoder to reduce the evolution complexity of the physical states via shared dynamic fields. 

For a scene containing $N$ Gaussians, if the state of each Gaussian is represented by a $G$-dimensional vector, directly evolving all $N\times G$ parameters independently is computationally prohibitive for large $N$. We observe that in physical particle systems such as the Material Point Method~\cite{mpm}, each particle retains static properties (\textit{e.g.}, mass, material), while its temporal evolution is driven by \textbf{shared} dynamic fields (\textit{e.g.}, gravitational fields) across particles. 

Following this idea, we factorize the state of the Gaussians into two components: 1) a set of $N$ particle-level static features $L \in \mathbb{R}^{N \times S}$, and 2) a set of $F$ system-level dynamic fields $g_t \in \mathbb{R}^{F \times G}$ that are shared across all Gaussians. 
The \textbf{\textit{physical state}} set $Z_t$ for all Gaussians is defined as:
\begin{equation}
Z_t = \mathrm{concat}[L, \textbf{1}_N\otimes g_t^\mathrm{flat}],\; Z_t\in \mathbb{R}^{N\times (S+F\cdot G)}.
\label{lgencoding}
\end{equation}
Here, $g_t^\mathrm{flat} \in \mathbb{R}^{1\times (F \cdot G)}$ is the flattened dynamic fields, $\mathbf{1}_N \in \mathbb{R}^{N\times1}$ is a vector of ones, and the Kronecker product $\mathbf{1}_N\otimes g_t^\mathrm{flat}$ broadcasts the global field to all $N$ particles. By evolving only $F$ dynamic fields rather than all $N$ particle states, we reduce the computational complexity of the dynamics evolver from $\mathcal{O}(NG)$ to $\mathcal{O}(FG)$, where $F \ll N$.

To obtain the initial physical state $Z_0$, we design a Dynamics Latent Space Encoder $f_{\mathrm{encoder}}$ as shown in Fig.~\ref{overview}(c). Gaussian features are first transformed into static features $L$ via linear layers. To handle different numbers of Gaussian particles, neighborhood Gaussian patches are constructed by farthest point sampling and $k$-nearest neighbors within a Mini-PointNet++ module~\cite{qi2017pointnet++}. Finally, we employ cross-attention blocks with $F$ learnable queries and self-attention to hierarchically aggregate $F$ initial dynamic fields $g_0$. Thus, the initial physical state set for the Gaussians $Z_0$ is encoded by $f_{\mathrm{encoder}}$ according to Eq.~\ref{lgencoding} as:
\begin{equation}
    f_{\mathrm{encoder}}(P) = Z_0.
\end{equation}
Here $P$ is the canonical Gaussian set, and each $p\in P$ is a Gaussian kernel with its parameters $\{x,sh,r,s,\alpha\}$.

\begin{figure}
    \centering
    \includegraphics[width=1.0\linewidth]{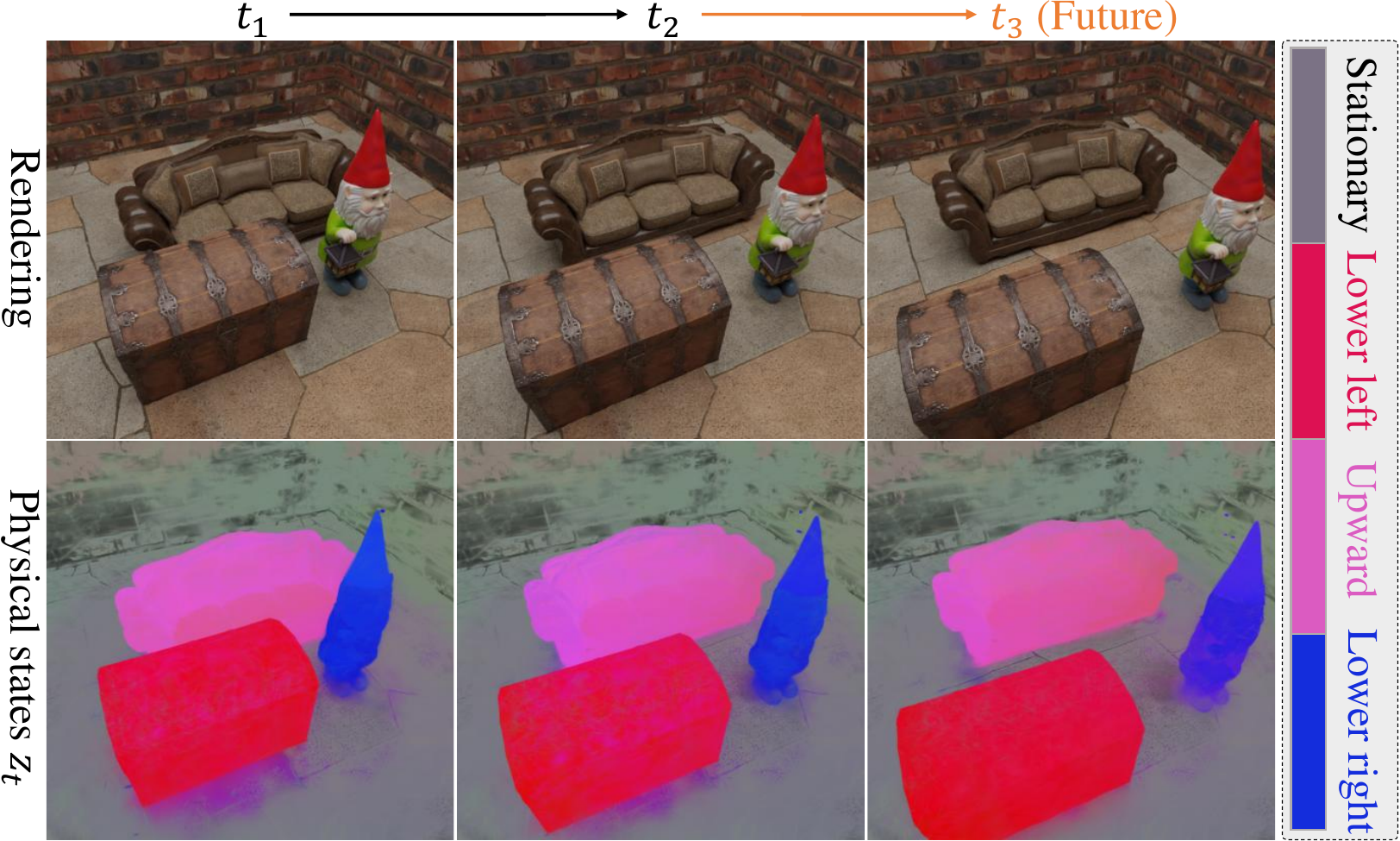}
    \caption{Rendering and visualization of physical states. Motion-similar Gaussians exhibit comparable features, and their physical states change stably over time.}
    \label{vis_z}
    \vspace{-0.4cm}
\end{figure}

The $F$ dynamic fields can be interpreted as a basis decomposition of the dynamics, where each field captures a distinct motion mode. To further analyze the role of physical states, we visualize them in RGB as shown in Fig.~\ref{vis_z}. It shows that: 1) Dynamic Coherence: The physical states $Z_t$ are not arbitrary but are strongly correlated with the motion patterns of the scene; 2) Temporal Stability: These stable and smooth physical states reflect a continuous and consistent physical evolution learned by the model.

\textbf{Neural ODEs Dynamics Evolver.}
Once the 3D Gaussians are encoded into the initial physical states, the core problem becomes evolving this state in a physically plausible manner. Rather than relying on low-order dynamics methods such as a velocity field, we propose a Neural ODEs Dynamics Evolver that models higher-order dynamics. 

Although the factorized encoding balances efficiency and representational capacity, a key challenge remains: in real motion, physical fields are often governed by high-order differential equations (\textit{e.g.}, Newton’s laws). Discrete-step models such as RNNs or MLPs, which model discrete updates, may struggle to capture continuous-time, high-order dynamics without careful design~\cite{ode}. 

To model such dynamics, we recall that any $n$-th order differential equation
\begin{equation}
\frac{d^n x}{dt^n} = f\left(x, \frac{dx}{dt}, \dots, \frac{d^{n-1}x}{dt^{n-1}}, t\right)
\end{equation}
can be equivalently written as a \emph{first-order} system $F$ by augmenting the state with its successive derivatives:
\begin{equation}
X = \big(x, \tfrac{dx}{dt}, \dots, \tfrac{d^{n-1}x}{dt^{n-1}}\big), \quad
\frac{dX}{dt} = F(X,t).
\label{firstorder}
\end{equation}
This formulation shows that a first-order system defined on an augmented latent space is expressive enough to represent any high-order dynamics.

Based on this principle, we use our dynamic fields $g_t$ to represent this augmented state and use a Neural ODE-based network $f_{\mathrm{evolver}}$ to model the high-order dynamics as visualized in Fig.~\ref{overview}(d):
\begin{equation}
\frac{dg_t}{dt} = f_{\mathrm{evolver}}(g_t, t).
\label{eq:node_deriv}
\end{equation}
The dynamic fields at any future time $t+\delta t$ are then obtained by integrating $f_{\mathrm{evolver}}$ using a numerical ODE solver:
\begin{equation} 
\begin{aligned} 
g_{t+\delta t} 
&= g_t + \int_{t}^{t+\delta t} f_{\mathrm{evolver}}(g_\tau, \tau) d\tau \\
&= \mathrm{ODESolver}(f_{\mathrm{evolver}}, g_t, t, t + \delta t).
\label{g02gt} 
\end{aligned} 
\end{equation}
Here, we use the common Runge–Kutta (RK4)~\cite{kutta1901beitrag} method as the numerical ODE solver.

Instead of memorizing a trajectory, our $f_{\mathrm{evolver}}$ learns to model the local high-order derivatives governing the dynamic fields. As a result, the learned physical laws can be explicitly followed by integrating $f_{\mathrm{evolver}}$, enabling stable and physically consistent extrapolation.

\textbf{Gaussian Kernel Space Decoder.} 
Following existing methods~\cite{li2025trace}, we decompose each Gaussian kernel’s motion into translation and rotation by Rodrigues’ rotation formula~\cite{rodrigues1840}. Specifically, the decoder predicts a translation vector $T$, a motion rotation vector $R$, and deformation terms $\{\delta r, \delta s\}$, which are used to update the parameters of Gaussian kernels $\{x, r, s\}$ as:
\begin{equation}
x_t = x + \mathrm{Rod}(R)x + T,\; r_t = r \circ \delta r,\; s_t = s + \delta s.
\label{gsdecode}
\end{equation}
Here $\mathrm{Rod}(R)$ is the rotation matrix generated from $R$ via Rodrigues’ formula~\cite{rodrigues1840}, and $\circ$ is the quaternion product.

We implement the decoder using a multi-head MLP $f_{\mathrm{decoder}}$, which maps each particle’s physical state $z_t \in Z_t$ to its corresponding deformation parameters:
\begin{equation}
f_{\mathrm{decoder}}(z_t) = \{T, R, \delta r, \delta s\}.
\label{eq:decoder}
\end{equation}
The final Gaussian parameters $p_t$ are obtained by applying these deformations to the original parameters $p$ by Eq.~\ref{gsdecode}.

This motion factorization enables the model to learn meaningful physical components of motion.

\subsection{Optimization}
\label{optimize}
Following~\cite{deformable3d}, we jointly optimize the Gaussian kernels and the ParticleGS network using $\mathcal{L}_1$ loss and D-SSIM loss:
\begin{equation}
    \mathcal{L}= \mathcal{L}_1+ \mathcal{L}_{D\text{-}SSIM}.
\end{equation}

\textbf{Progressive Training.} Directly optimizing both geometry and the Neural ODE-based deformation model from video supervision can lead to unstable training, as the ODE solver may propagate errors in early steps. To mitigate this, we adopt a progressive training schedule that gradually introduces complexity: 1) Geometry warm-up: freeze ParticleGS and optimize the Gaussian kernels at $t=0$; 2) Dynamics warm-up: freeze the Gaussian kernels and gradually expand the temporal window, allowing ParticleGS to learn deformation trends; 3) Joint optimization: optimize both the Gaussian kernels and ParticleGS.

\textbf{Dynamic Gaussian Optimization.} Following standard 3DGS~\cite{3dgs} and the existing dynamic 3DGS methods~\cite{4dgs,deformable3d,li2025freegave}, we periodically apply densification and pruning, meaning the total Gaussian count $N$ changes during training. For static feature $L$, the Encoder and Decoder MLPs treat the $N$ particles as a batch, operating on each independently as in existing methods. For initial dynamic fields $g_0$, we use a Mini-PointNet++ module to form a fixed number of neighborhood patches and use the Cross-Attention module to aggregate $F$ dynamic fields, thus it maps a variable-sized input $N$ to a fixed-size output $F$.

\textbf{Neighborhood Regularization.} To ensure the encoder is robust to different local particle structures resulting from densification, we propose an online neighborhood regularization. Instead of computing the neighborhood graph using Mini-PointNet++ only once, we periodically regenerate the neighborhood patches during training. This acts as a data augmentation, enhancing the robustness of the encoder for extracting the initial dynamic fields $g_0$.

See Appendix for more implementation details.

\vspace{-5pt}
\section{Experiments}
\begin{table*}
  \caption{Quantitative results on the Dynamic Objects and Dynamic Indoor Scene dataset, rendered at the source resolution. \textbf{Bold} and \underline{underline} indicate the best and second best performance.}
  \label{tab:nvfi}
  \centering
  \resizebox{\linewidth}{!}{
  \begin{tabular}{lcccccccccccc}
    \toprule
    \multirow{3}{*}{Method} 
      & \multicolumn{6}{c}{Dynamic Object Dataset (Synthetic)} 
      & \multicolumn{6}{c}{Dynamic Indoor Scene Dataset (Synthetic)} \\
    \cmidrule(lr){2-7} \cmidrule(lr){8-13}
      & \multicolumn{3}{c}{Reconstruction} & \multicolumn{3}{c}{Extrapolation} 
      & \multicolumn{3}{c}{Reconstruction} & \multicolumn{3}{c}{Extrapolation} \\
    \cmidrule(r){2-4} \cmidrule(r){5-7} \cmidrule(r){8-10} \cmidrule(r){11-13}
      & PSNR$\uparrow$ & SSIM$\uparrow$ & LPIPS$\downarrow$ & PSNR$\uparrow$ & SSIM$\uparrow$ & LPIPS$\downarrow$  
      & PSNR$\uparrow$ & SSIM$\uparrow$ & LPIPS$\downarrow$ & PSNR$\uparrow$ & SSIM$\uparrow$ & LPIPS$\downarrow$ \\
    \midrule
    HexPlane~\cite{hexplane} & 21.585  & 0.910  & 0.125  & 22.418  & 0.934  & 0.078  & 17.978  & 0.482  & 0.626  & 23.192  & 0.671  & 0.474  \\ 
    TiNeuVox~\cite{TiVox} & 20.278  & 0.927  & 0.104  & 19.513  & 0.935  & 0.080  & \underline{23.706}  & \underline{0.681}  & \underline{0.373}  & 21.111  & 0.714  & 0.333 \\ 
    DeformGS~\cite{deformable3d} & 37.376  & 0.964  & 0.034  & 26.163  & 0.946  & 0.038 & 20.020  & 0.595  & 0.445  & 21.982  & 0.769  & 0.233  \\ 
    Grid4D~\cite{grid4d} & 37.583  & 0.961  & 0.034  & 27.674  & 0.949  & 0.080  & 20.548  & 0.590  & 0.442  & 20.633  & 0.736  & 0.284  \\ 
    NVFi~\cite{nvfi} & 29.390  & 0.969  & 0.047  & 26.821  & 0.968  & 0.047  & 17.117  & 0.494  & 0.669  & 23.716  & 0.672  & 0.502   \\ 
    GSPrediction~\cite{gsp} & 36.408  & 0.978  & 0.033  & 23.154  & 0.932  & 0.081  & 16.154  & 0.585  & 0.600  & 20.017  & 0.689  & 0.343    \\ 
    TRACE~\cite{li2025trace} & 38.006  & \underline{0.991}  & \underline{0.011}  & 33.356  & \underline{0.983}  & \underline{0.013} & 22.846  & 0.642  & 0.377  & \underline{29.476}  & 0.849  & \underline{0.205} \\
    FreeGave~\cite{li2025freegave}  & \underline{38.637}  & \textbf{0.992}  & \underline{0.011}  & \underline{33.632}  & \underline{0.983}  & \textbf{0.012} & 19.681  & 0.589  & 0.453  & 28.983  & \underline{0.854}  & 0.212 \\
    \midrule
    \textbf{ParticleGS} & \textbf{39.782}  & \textbf{0.992}  &\textbf{ 0.009}  & \textbf{36.471}  & \textbf{0.984}  & \textbf{0.012} & \textbf{25.504}  & \textbf{0.748}  & \textbf{0.291}  & \textbf{31.103}  & \textbf{0.892}  & \textbf{0.133}    \\ 
    \bottomrule
  \end{tabular}
  }

  \vspace{0.2cm}

  \caption{Quantitative results on the Dynamic Multipart dataset and FreeGave-GoPro dataset, rendered at the source resolution. \textbf{Bold} and \underline{underline} indicate the best and second best performance.}
  \label{tab:gopro}

  \centering
  \resizebox{\linewidth}{!}{
  \begin{tabular}{lcccccccccccc}
    \toprule
    \multirow{3}{*}{Method} 
      & \multicolumn{6}{c}{Dynamic Multipart Dataset (Synthetic)} 
      & \multicolumn{6}{c}{FreeGave-GoPro Dataset (Real World)} \\
    \cmidrule(lr){2-7} \cmidrule(lr){8-13}
      & \multicolumn{3}{c}{Reconstruction} & \multicolumn{3}{c}{Extrapolation} 
      & \multicolumn{3}{c}{Reconstruction} & \multicolumn{3}{c}{Extrapolation} \\
    \cmidrule(r){2-4} \cmidrule(r){5-7} \cmidrule(r){8-10} \cmidrule(r){11-13}
      & PSNR$\uparrow$ & SSIM$\uparrow$ & LPIPS$\downarrow$ & PSNR$\uparrow$ & SSIM$\uparrow$ & LPIPS$\downarrow$  
      & PSNR$\uparrow$ & SSIM$\uparrow$ & LPIPS$\downarrow$ & PSNR$\uparrow$ & SSIM$\uparrow$ & LPIPS$\downarrow$ \\
    \midrule
    DeformGS~\cite{deformable3d} & \underline{38.430}  & \underline{0.988} & \textbf{0.019}  & 27.989  & 0.942  & 0.045 & \underline{20.106} & 0.832 & 0.212 & 21.665 & 0.858 & 0.195 \\
    TRACE~\cite{li2025trace} & 36.269  & \textbf{0.990}  & 0.021  & 33.458  & \underline{0.984} & \underline{0.015} & 20.103  & \underline{0.838}  & \textbf{0.205}  & 25.915  & 0.892  & 0.163 \\
    FreeGave~\cite{li2025freegave} & 36.142  & \textbf{0.990}  & \underline{0.020}  & \underline{33.526} & 0.981  & 0.019 & 19.953 & \underline{0.838} & \underline{0.206} & \underline{26.510} & \underline{0.896} & \underline{0.160} \\
    \midrule
    \textbf{ParticleGS} & \textbf{38.608} & \textbf{0.990} & \textbf{0.019} & \textbf{36.136} & \textbf{0.985} & \textbf{0.014} & \textbf{20.116} & \textbf{0.839}& \textbf{0.205}& \textbf{26.789} & \textbf{0.903} & \textbf{0.150} \\ 
    \bottomrule
  \end{tabular}
  
  }
  \vspace{-0.4cm}
\end{table*}

\begin{table}
    \centering
    \caption{Quantitative results of FPS$\uparrow$. \textbf{Bold} and \underline{underline} indicate the best and second best performance.}

    \resizebox{\linewidth}{!}{
    \begin{tabular}{lcc}
    \toprule
    Method & Dynamic Object Dataset & Dynamic Indoor Scene Dataset \\
    \midrule
    TRACE~\cite{li2025trace} & \textbf{51.041}  & \textbf{39.590}  \\
    FreeGave~\cite{li2025freegave} & 32.303  & 32.129  \\
    \midrule
    \textbf{ParticleGS} & \underline{44.293}  & \underline{37.109}  \\
    \bottomrule
    \end{tabular}
    }
    \label{fps}
    \vspace{-0.5cm}
\end{table}

\subsection{Experimental Setup}
\textbf{Datasets.} This work's primary focus is \textbf{temporal motion extrapolation}, rather than only spatial novel view synthesis of observed frames. To this end, we follow the experimental setup of the most closely related works, NVFi~\cite{nvfi} and FreeGave~\cite{li2025freegave}, and conduct our primary evaluations on the Dynamic Object Dataset and Dynamic Indoor Scene Dataset, which are specifically designed for extrapolation tasks~\cite{nvfi}. Furthermore, we evaluate on Dynamic Multipart~\cite{li2025trace} and the challenging real-world dataset FreeGave-GoPro~\cite{li2025freegave}. A brief description of the datasets is as follows: 1) \textit{Dynamic Object Dataset}: Contains 6 objects exhibiting a wide range of motions, including complex nonlinear motions of deformable animals such as flying and swimming; 2) \textit{Dynamic Indoor Scene Dataset}: Consists of 4 scenes with multiple objects undergoing compositional motions; 3) \textit{Dynamic Multipart}: Includes four objects with distinct multi-part motions. 4) \textit{FreeGave-GoPro}: Covers 6 human interactions with objects in the real world. Following~\cite{nvfi, li2025freegave}, all datasets are divided into subsets for \textit{reconstruction} and \textit{extrapolation} tasks by training on the first 75\% of video frames and test on the remaining 25\%.

\textbf{Baselines.} The key distinction of ParticleGS lies in its physics-based modeling and its ability to learn high-order dynamics by Neural ODEs. To evaluate the distinctions of ParticleGS over baselines, we select three categories of baselines: 1) Representative methods based on time-deformation modeling: HexPlane~\cite{hexplane}, TiNeuVox~\cite{TiVox}, DeformGS~\cite{deformable3d}, and Grid4D~\cite{grid4d} are representative reconstruction methods; 2) Methods that embed physical information for motion extrapolation: NVFi~\cite{nvfi} is a PINN-based method that embeds the Navier-Stokes equations, GaussianPrediction~\cite{gsp} is a GCN-based method that embeds time-invariant rigid neighborhood relations; 3) State-of-the-art velocity field-based method for extrapolation: FreeGave~\cite{li2025freegave} and TRACE~\cite{li2025trace} learn motion via divergence-free velocity fields. All methods are run on a single RTX 4090 with 24GB GPU memory, except for an ablation study run on an A6000 with 48GB.

\textbf{Metrics.} We follow existing dynamic 3D extrapolation works~\cite{nvfi,li2025freegave,li2025trace} and use the standard \textbf{PSNR}, \textbf{SSIM}, \textbf{LPIPS}, and \textbf{FPS} metrics to evaluate the dynamic 3D reconstruction and future frame extrapolation tasks.

\begin{figure*}
    \centering

    \includegraphics[width=1.0\linewidth]{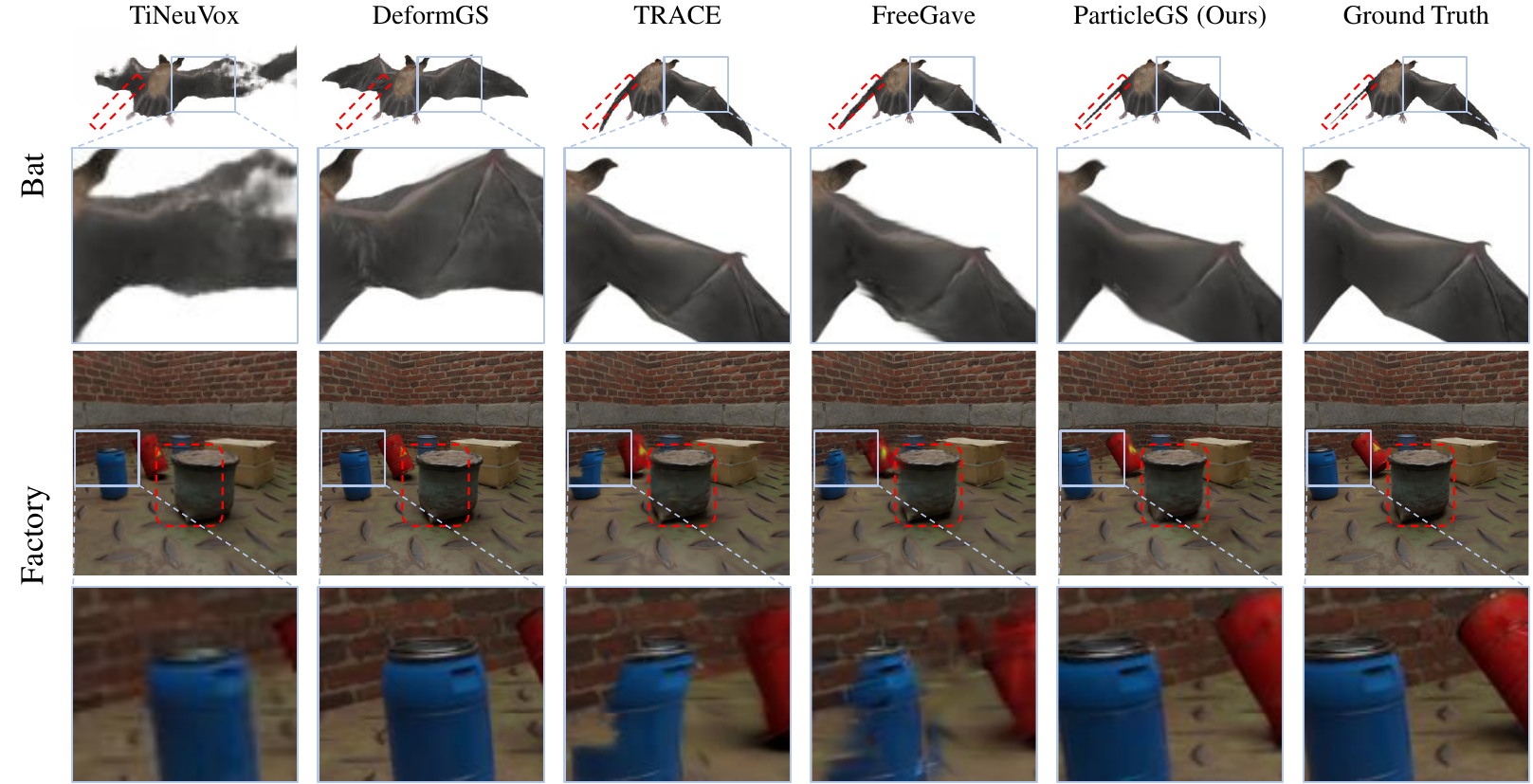}
    \caption{Qualitative results of extrapolation at a novel view and a future time on the Dynamic Object Dataset (Bat) and the Dynamic Indoor Scene Dataset (Factory).  Red boxes indicate ground truth locations, and blue boxes highlight details.}
    \label{exp_fig1}

    \vspace{0.3cm}

    \centering
    \includegraphics[width=1.0\linewidth]{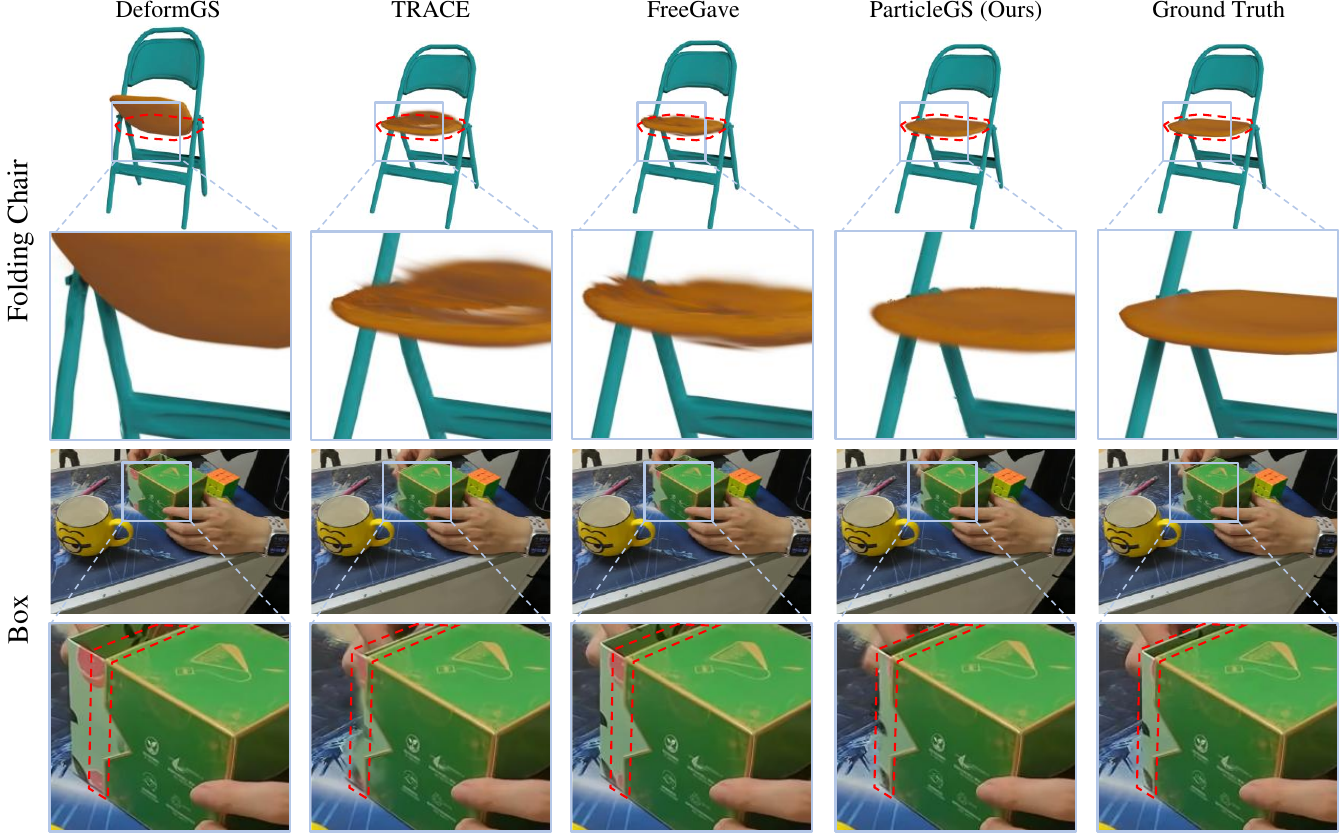}
    \caption{Qualitative results of extrapolation at a novel view and a future time on the Dynamic Multipart Dataset (Folding Chair) and the FreeGave-GoPro Dataset (Box). Red boxes indicate ground truth locations, and blue boxes highlight details.}
    \label{exp_fig2}

\end{figure*}

\subsection{Comparisons with Prior Work}
\begin{table*}
    \centering
    \caption{Results of extrapolation ablation studies. The method marked * indicates it is run on an A6000 with 48GB.}

    \resizebox{\linewidth}{!}{
    \begin{tabular}{lccccccccccc}
    \toprule
    \multirow{2}[0]{*}{ } & & & & & & \multicolumn{3}{c}{Dynamic Object Dataset} & \multicolumn{3}{c}{Dynamic Indoor Scene Dataset} \\
    \cmidrule(lr){7-9} \cmidrule(lr){10-12}
     Method & Encoder & Evolver & Decoder & Optimize & $F$ & PSNR$\uparrow$ & SSIM$\uparrow$ & LPIPS$\downarrow$ & PSNR$\uparrow$ & SSIM$\uparrow$ & LPIPS$\downarrow$ \\
    \midrule
    (1) *   &  w/o FE & full  & full  & full  & \usym{2717}  & 36.391  & 0.982  & \textbf{0.012}  & -     & -     & - \\
    (2.1) & full  & full  & full  & full     & 1  & 35.206  & 0.982  & \textbf{0.012}  & 28.026  & 0.883  & 0.174  \\
    (2.2) & full  & full  & full  & full     & 4  & 36.236  & \underline{0.983}  & \textbf{0.012}  & 30.977  & \underline{0.890}  & \underline{0.135}  \\
    (2.3) & full  & full  & full  & full    & 16  & \textbf{36.552}  & \textbf{0.984}  & \textbf{0.012}  & \underline{31.065}  & \textbf{0.892}  & \textbf{0.133}  \\
    (3)   & full  & w/o ODEs & full  & full   & 8  & 34.551  & 0.981  & 0.015  & 27.455  & 0.861  & 0.179  \\
    (4)   & full  & full  & w/o PD & full    & 8  & 35.343  & 0.982  & 0.014  & 28.653  & 0.879  & 0.198  \\
    (5)   & full  & full  & full  & w/o PT   & 8 & 35.438  & 0.979  & 0.015  & 29.168  & 0.883  & 0.218  \\
    (6)   & full  & full  & full  & w/o NR   & 8 & 35.986  & 0.981  & 0.015  & 30.232  & \underline{0.890}  & 0.161  \\
    \midrule
    \textbf{ParticleGS}   & full  & full  & full  & full     & 8  & \underline{36.471}  & \textbf{0.984}  & \textbf{0.012}  & \textbf{31.103}  & \textbf{0.892}  & \textbf{0.133}  \\
    \bottomrule
    \end{tabular}
    \label{tab:ablation}
    }
    \vspace{-0.4cm}
\end{table*}
We first compare all the methods on the Dynamic Object Dataset and the Dynamic Indoor Scene Dataset. Then, we compare 3 representative methods on the Dynamic Multipart and the challenging GoPro Dataset. 

\textbf{Quantitative results of performance \& Analysis.} Tables \ref{tab:nvfi} and \ref{tab:gopro} show that our method achieves the best performance in both reconstruction and extrapolation tasks. 

On the extrapolation task, our method exhibits significant advantages: 1) Compared to existing time-conditioned methods and approaches that inject physical priors, our method achieves a substantial PSNR improvement of over 5 dB across all datasets. This result suggests that our model successfully learns the underlying physical laws from observational data and follows them during extrapolation. 2) Furthermore, ParticleGS outperforms velocity field-based extrapolation methods, such as TRACE and FreeGave, by an average PSNR margin of nearly 2.5 dB on synthetic datasets. This performance advantage extends to the challenging real-world FreeGave-GoPro Dataset. This indicates that our method can capture complex high-order dynamics more accurately than such low-order velocity field-based methods, leading to superior predictive accuracy.

For the reconstruction task, our method also achieves the best performance across all datasets. This demonstrates that our modeling framework is superior to or comparable to existing reconstruction methods.

\textbf{Quantitative results of speed \& Analysis.} Tab.~\ref{fps} shows that our method achieves a rendering speed comparable to velocity field–based approaches. Although we incorporate Neural ODEs, our physical state formulation requires only a single forward evaluation of the $F$ dynamic fields, rather than computing the velocity of all $N$ Gaussians. This validates the efficiency of our physical state design.

\textbf{Qualitative comparisons of visualization results.}
As shown in Fig. \ref{exp_fig1} and Fig. \ref{exp_fig2}, our method demonstrates superior visual fidelity and physical consistency compared to existing approaches: 1) For the representative time-conditioned methods TiNeuVox and DeformGS, the predicted results highlighted by red boxes exhibit clear spatial misalignments, whereas our method accurately aligns with the ground-truth locations. This demonstrates that our approach can extrapolate more physically consistent Gaussian deformations. 2) In the Factory and Folding Chair datasets, velocity field-based TRACE and FreeGave yield blurrier results than ParticleGS, and in the Box real-world dataset, their motion predictions appear noticeably misaligned. In contrast, ParticleGS delivers clearer and more physically plausible results, validating the effectiveness of our high-order dynamic modeling enabled by Neural ODEs.

\subsection{Ablation Studies and Analysis}
\label{ablation}
To investigate the impact of different components of our framework, we conducted the following ablation studies.

\textbf{(1) Effect of Factorized Encoding (FE) in Eq.~\ref{lgencoding}.} Instead of decomposing the latent state into static properties $l$ and dynamic fields $g_t$, we encode each Gaussian with a feature of the same dimensionality. This tests whether the encoding strategy is essential for balancing representational capacity and computational efficiency. 

\textbf{(2) Number of Dynamic Fields $F$ in Eq.~\ref{lgencoding}.} We vary $F \in \{1, 4, 8, 16\}$ and use $F=8$ in the comparative experiment. This ablation validates our claim that $F$ fields suffice to capture diverse yet structured dynamics. 

\textbf{(3) Neural ODEs Dynamics Evolver.} We replace Neural ODEs with MLPs with the same number of parameters. This aims to validate the benefits of dynamics modeling in learning physical dynamics.

\textbf{(4) Effect of Physical Decoding (PD) in Eq.~\ref{gsdecode}.} We remove Rodrigues’ rotation formula~\cite{rodrigues1840} to update the motion. This tests whether physically meaningful transformations help maintain physically plausible motion. 

\textbf{(5) Removing Progressive Training (PT).} We disable the deformation warm-up to align with existing methods. This evaluates the importance of progressive training in disentangling geometry and dynamics. 

\textbf{(6) Removing Neighborhood Regularization (NR).} We compute the neighborhood graph once and cache it throughout training. This tests whether online updating neighborhood partitions enhances model robustness.

\textbf{Results \& Analysis:} Tab. \ref{tab:ablation} shows that: \textbf{(1)} Removing the factorized encoding significantly increases GPU memory consumption even on object-level 3D scenes, while bringing no performance improvement. Since motion exhibits correlation, assigning an independent dynamic state to each particle not only greatly increases computational complexity but also slightly degrades performance. This demonstrates that our factorized encoding is crucial for striking a balance between computational efficiency and representational capacity. \textbf{(2)} Using too few dynamic feature fields ($F=1$) leads to a noticeable performance drop, while increasing $F$ beyond a moderate number ($F=16$) yields limited gains. This suggests that a sufficient yet compact set of dynamic feature fields is sufficient to capture the regular motion patterns. \textbf{(3)} Replacing Neural ODE-based Dynamics Evolver with MLPs causes the largest performance degradation. Since the dynamics typically follow high-order, continuous differential equations, Neural ODEs are inherently more suitable for modeling such dynamics. This confirms the necessity of introducing Neural ODEs into our framework. \textbf{(4)} Removing the physically meaningful decoding formulation noticeably reduces extrapolation accuracy, indicating that the physically consistent decoding process helps the model learn realistic dynamic behaviors. \textbf{(5–6)} Excluding either the progressive training or the neighborhood regularization leads to lower performance, verifying that these strategies improve the robustness of the proposed ParticleGS model.

\section{Conclusion} We introduce ParticleGS, a dynamic 3DGS method that learns neural dynamics directly from video observations. By modeling physical states and learning state transition laws, ParticleGS ensures stable extrapolation. Experimental results demonstrate that ParticleGS outperforms existing baselines across multiple benchmarks.

\textbf{Limitations.} Similar to other methods, ParticleGS currently cannot predict motions that deviate from the physical dynamics observed during training (e.g., sudden object breakage), as it has not learned to model such behaviors.

{
    \small
    \bibliographystyle{ieeenat_fullname}
    \bibliography{main}
}

\clearpage
\setcounter{page}{1}
\maketitlesupplementary

The appendix includes:
\begin{itemize}
\item A Toy Experiment: Learning Dynamics vs. Memorizing Trajectories.
\item Implementation Details.
\item Additional Qualitative Comparisons.
\item Full Quantitative Results.
\end{itemize}

\section{A Toy Experiment: Learning Dynamics vs. Memorizing Trajectories}
\label{sec:toy_exp}

\begin{figure}
    \centering
    \includegraphics[width=1.0\linewidth]{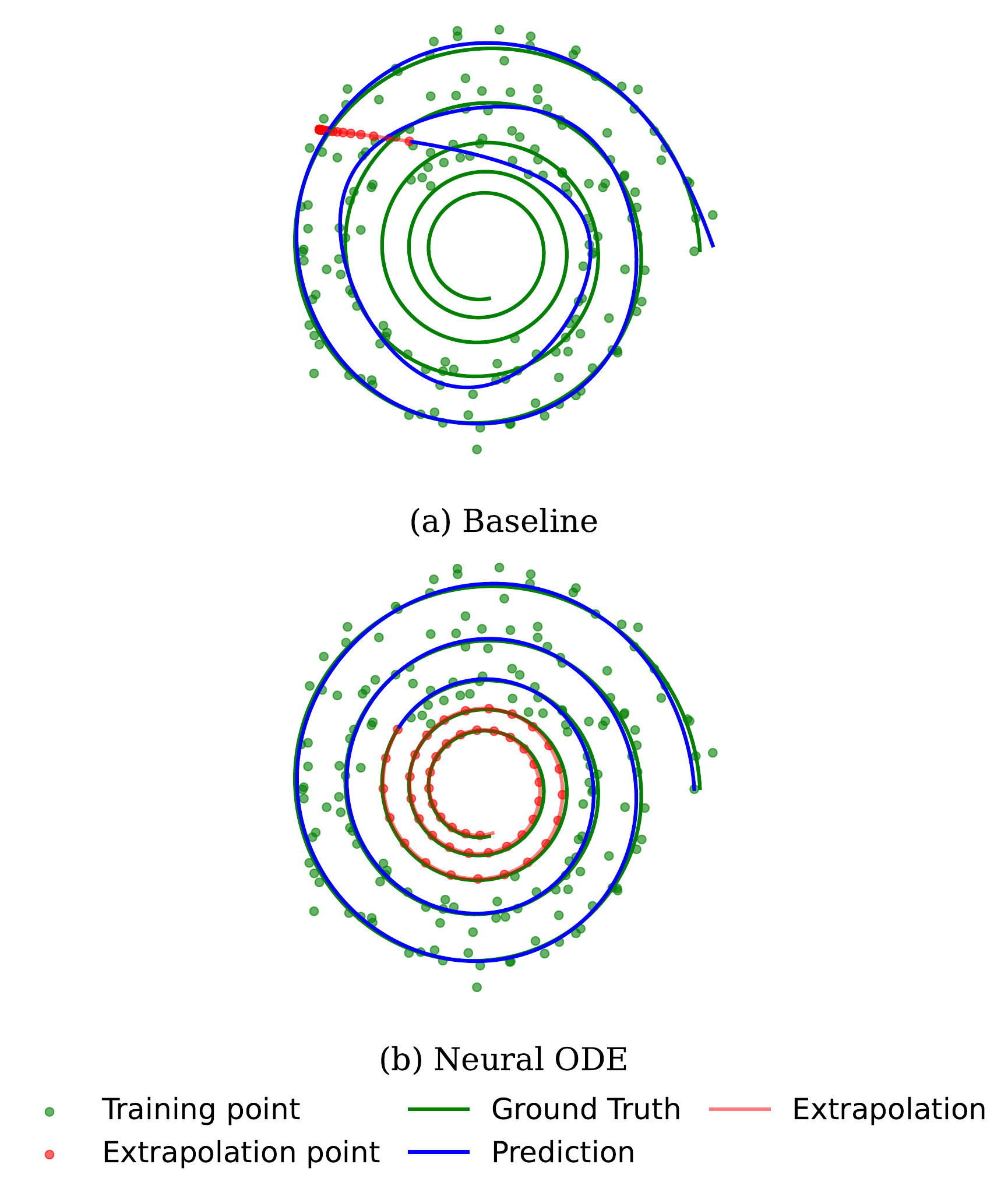}
    \caption{\textbf{Toy experiment on extrapolation capabilities.} We visualize the predicted trajectories on a 2D damped spiral. Green dots represent observed training data, while red dots indicate the unobserved future. \textbf{(a) Baseline:} The time-conditioned baseline fits the training data well but fails during extrapolation, producing a linear trajectory that deviates from the physical spiral. \textbf{(b) Neural ODE:} By modeling the underlying derivatives, the Neural ODE accurately captures the spiral dynamics, maintaining the correct trajectory even in the unobserved region.}
    \label{exp_toy}
    \vspace{-0.5cm}
\end{figure}

To intuitively understand the limitations of time-conditioned representations and the necessity of our physics-based approach, we conducted a toy experiment on a 2D damped harmonic oscillator system (a spiral trajectory), as shown in Fig.~\ref{exp_toy}. We trained two models on the initial segment of the spiral (corrupted by random noise, indicated by green dots) and tested their ability to predict the future evolution (indicated by red dots).

\textbf{Analysis of Time-Conditioned Models.}
As illustrated in Fig.~\ref{exp_toy}(a), the baseline method, which models position as a direct function of time ($x = \mathcal{F}(t)$), successfully interpolates the training points. However, it fails to extrapolate to the future. This is because the baseline essentially performs curve fitting on the time coordinate, memorizing the correlation between a specific timestamp $t$ and a position $x$ within the observed window. Once $t$ extends beyond this range, the network lacks the underlying physical context to predict the future, reverting to linear behavior that violates the system's inherent dynamics.

\textbf{Analysis of Neural ODEs.}
In contrast, Fig.~\ref{exp_toy}(b) demonstrates that the Neural ODE-based approach accurately predicts the future spiral motion. This success stems from the fact that Neural ODEs do not map time directly to position. Instead, they approximate the governing equation of the system: $\frac{dx}{dt} = f(x, t)$. While the position $x(t)$ changes constantly, the physical law $f$ governing the change remains consistent. By learning this derivative function from the observed past, the model can integrate the same law to generate physically plausible future states. This observation motivates ParticleGS to move from memorizing temporal deformations to learning the physical laws driving the system.

\section{Implementation Details}
\textbf{Input Representation.}
To capture the comprehensive state of each Gaussian particle, we construct a 14-dimensional feature vector as the input to our network. Specifically, this vector is a concatenation of the position $x \in \mathbb{R}^3$, rotation quaternion $r \in \mathbb{R}^4$, scaling $s \in \mathbb{R}^3$, opacity $\alpha \in \mathbb{R}^1$, and the 0-th order of the Spherical Harmonics coefficients $sh \in \mathbb{R}^3$. This compact representation enables the encoder to extract both geometric and photometric cues effectively.

\textbf{Dynamics Latent Space Encoder $f_{\text{encoder}}$:} 
The encoder is designed to extract particle-level static features and system-level dynamic fields. It consists of 4 stacked attention layers, each with a hidden dimension of 256, employing GELU as the activation function.
To handle the unstructured nature of 3D Gaussian clouds, we construct local Gaussian patches using Farthest Point Sampling (FPS) and $k$-nearest neighbors (k-NN).  We employ FPS to select representative anchor points from the Gaussian cloud. For synthetic datasets, we sample $N_k = 2048$ keypoints. For real-world datasets which typically contain more complex geometries and background noise, we increase the sampling density to $N_k = 4096$. For each keypoint, we query its $k=32$ nearest neighbors (k-NN) based on Euclidean distance to form a local patch. To adapt to the changing Gaussian distribution, we regenerate these neighborhood patches immediately after every densification step or at a fixed interval of 500 iterations.

This hierarchical design ensures efficient information aggregation across the scene.

\textbf{Neural ODE Dynamics Evolver $f_{\text{evolver}}$:} The evolver is the core component for modeling continuous-time dynamics. We implement $f_{\text{evolver}}$ using a network composed of two stacked 3-layer MLP blocks equipped with residual connections to facilitate gradient flow. All hidden layers maintain a dimension of 128 and use GELU activation. We utilize the Runge-Kutta 4 (RK4) method as the ODE solver to integrate the learned derivative function.

\textbf{Gaussian Kernel Space Decoder $f_{\text{decoder}}$:} We use a 6-layer MLP with a hidden dimension of 256 for all layers, using GELU as the activation function.

\textbf{Physical State Dimensions.}
 We set the dimension of the static features $L$ to 16 and the dimension of each dynamic field $G$ to 16. 

\textbf{Optimization.} We train our framework using the Adam optimizer and adopt a progressive training strategy. The total training iterations are set to 40,000 for synthetic datasets and 60,000 for real-world datasets. For 0-3k iterations: We adopt Geometry Warm-up and optimize only the static Gaussian parameters to obtain a reliable canonical geometry. For 3k-7k iterations: We adopt Dynamics Warm-up, where we freeze the canonical Gaussians and optimize the ParticleGS network with a gradually increasing temporal window. For 7k-40k/60k iterations: Both the canonical Gaussians and the neural networks are jointly optimized to refine fine-grained details until the end of training.

\section{Rodrigues' Rotation Formula}

In our framework, the decoder predicts a rotation vector $R \in \mathbb{R}^3$ (representing the axis-angle notation) to model the rotational dynamics. To apply this rotation to the 3D Gaussians, we convert $R$ into a valid rotation matrix $\mathrm{Rod}(R) \in \mathbb{R}^{3\times 3}$ using Rodrigues' rotation formula. Given the rotation vector $R = [r_x, r_y, r_z]^T$, the rotation angle $\theta$ is defined as its Euclidean norm:
\begin{equation}
\theta = \|R\|_2.
\end{equation}
If $\theta$ is non-zero, the normalized rotation axis $K = [k_x, k_y, k_z]^T$ is computed as $K = R / \theta$. We first define the skew-symmetric matrix $[K]_\times$ associated with the unit vector $K$:
\begin{equation}
[K]_\times =\begin{bmatrix}0 & -k_z & k_y 
\\ k_z & 0 & -k_x 
\\-k_y & k_x & 0\end{bmatrix}.
\end{equation}
Rodrigues' rotation formula then expresses the rotation matrix $\mathrm{Rod}(R)$ as:
\begin{equation}
\mathrm{Rod}(R) = I + \sin\theta [K]_\times + (1 - \cos\theta) [K]_\times^2,\end{equation}
where $I$ is the $3 \times 3$ identity matrix.

\section{Additional Qualitative Comparisons}
\label{sec:qualitative_results}

\begin{figure*}
    \centering
    \includegraphics[width=1.0\linewidth]{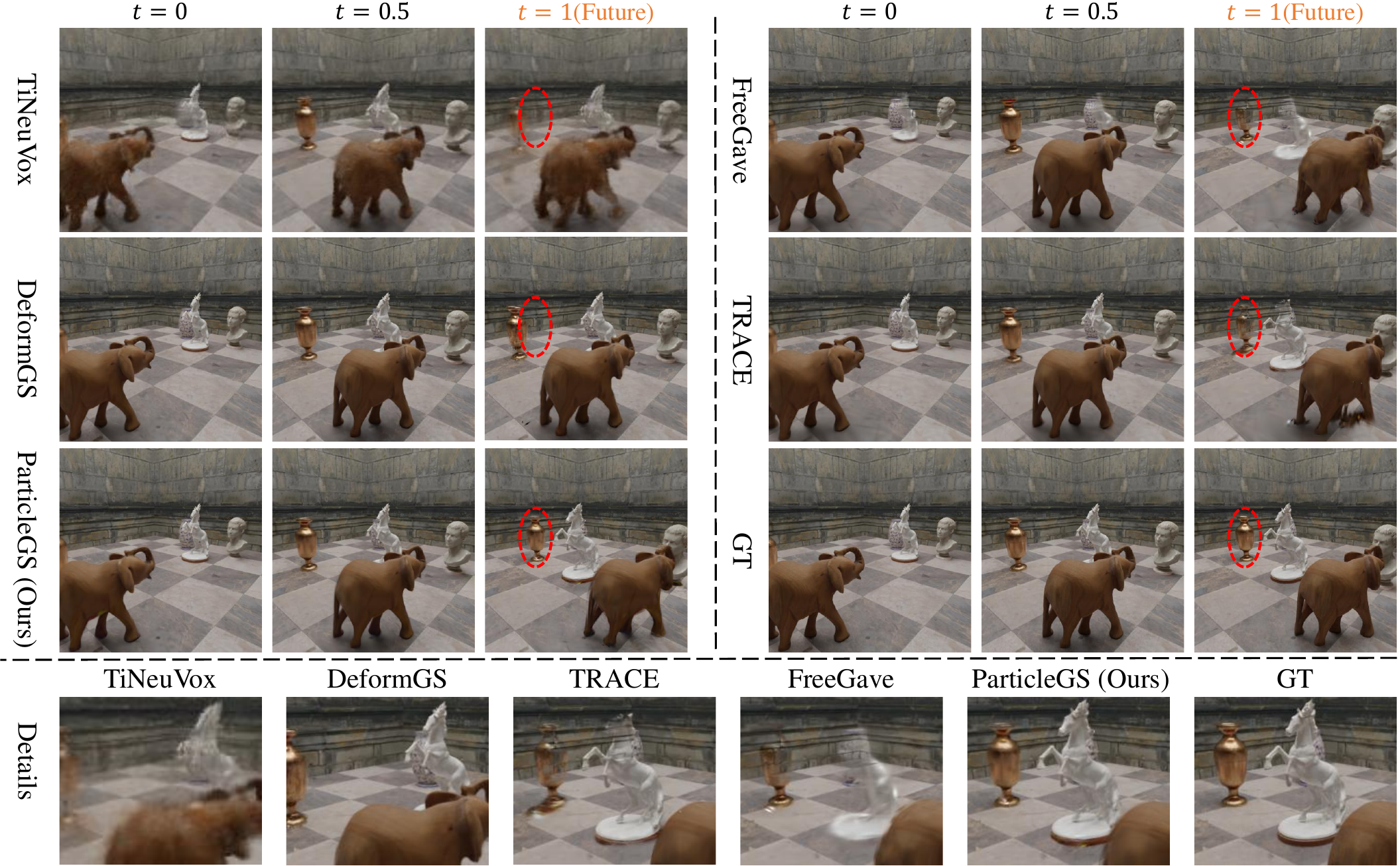}
    \caption{Additional qualitative comparisons on the Dynamic Indoor Scene Dataset.}
    \label{app_fig3}

    \vspace{0.3cm}
    
    \centering
    \includegraphics[width=1.0\linewidth]{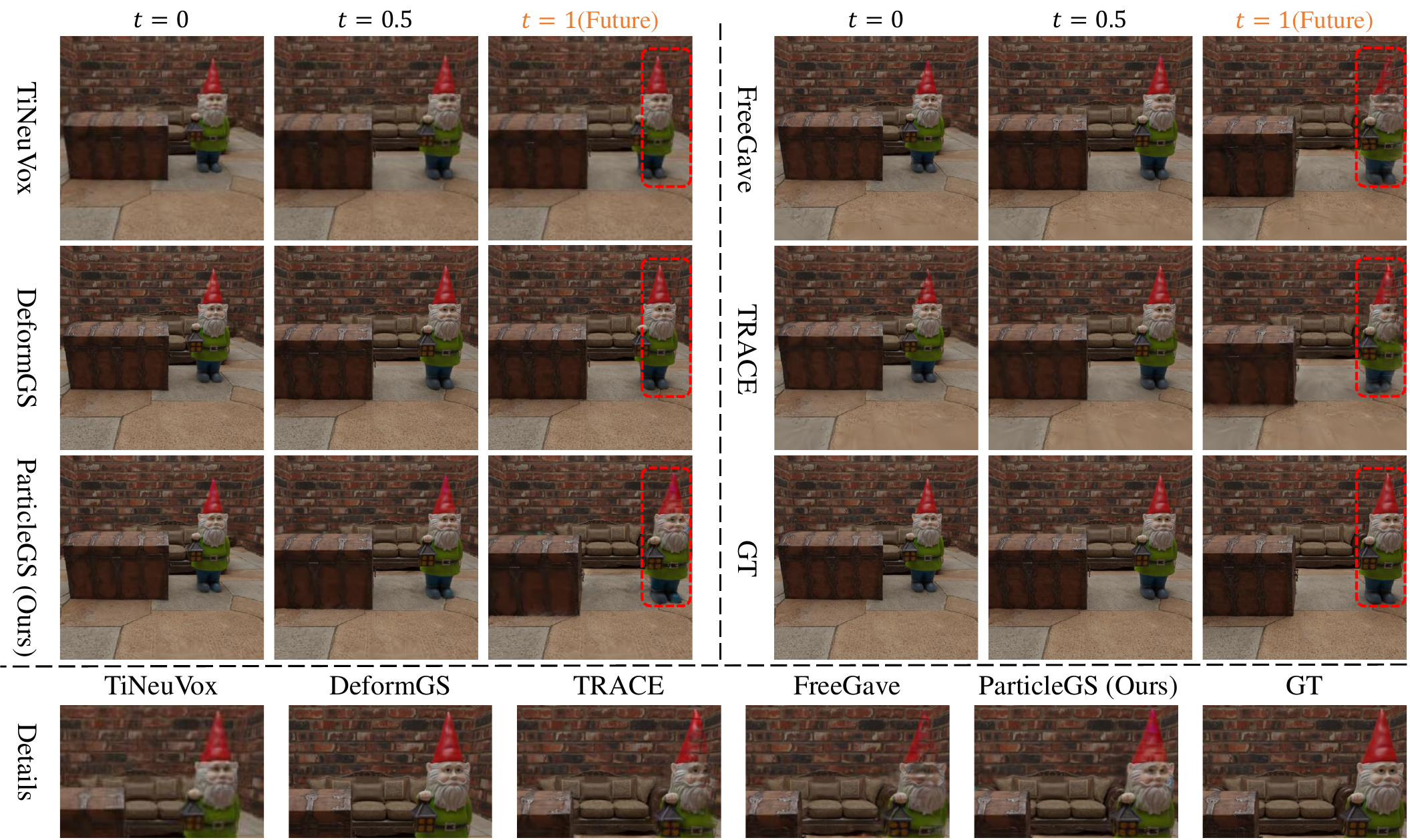}
    \caption{Additional qualitative comparisons on the Dynamic Indoor Scene Dataset.}
    \label{app_fig4}
\end{figure*}

\begin{figure*}
    \centering
    \includegraphics[width=1.0\linewidth]{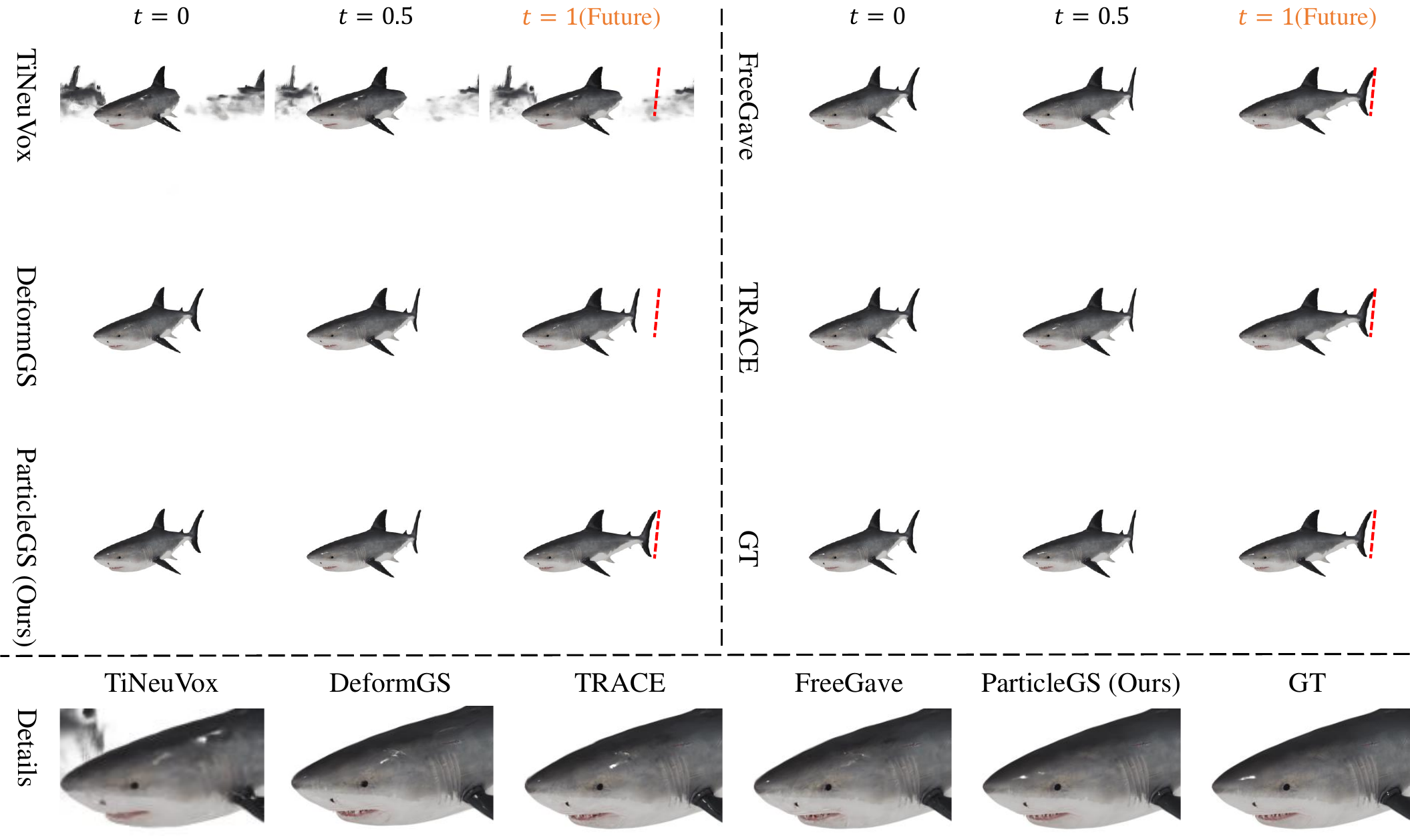}
    \caption{Additional qualitative comparisons on the Dynamic Object Dataset.}
    \label{app_fig1}

    \vspace{0.3cm}
    
    \centering
    \includegraphics[width=1.0\linewidth]{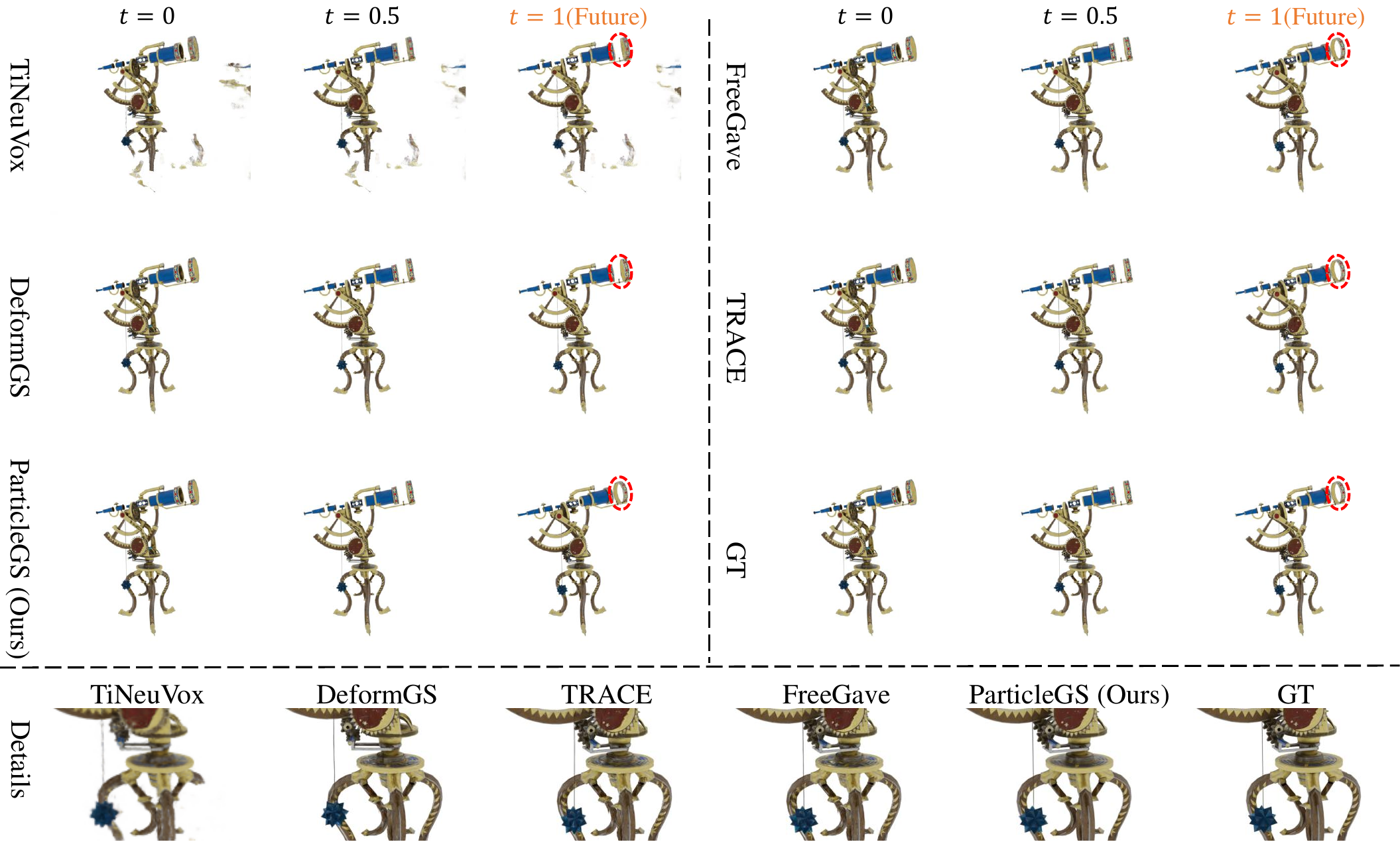}
    \caption{Additional qualitative comparisons on the Dynamic Object Dataset.}
    \label{app_fig2}
\end{figure*}

\begin{figure*}
    \centering
    \includegraphics[width=1.0\linewidth]{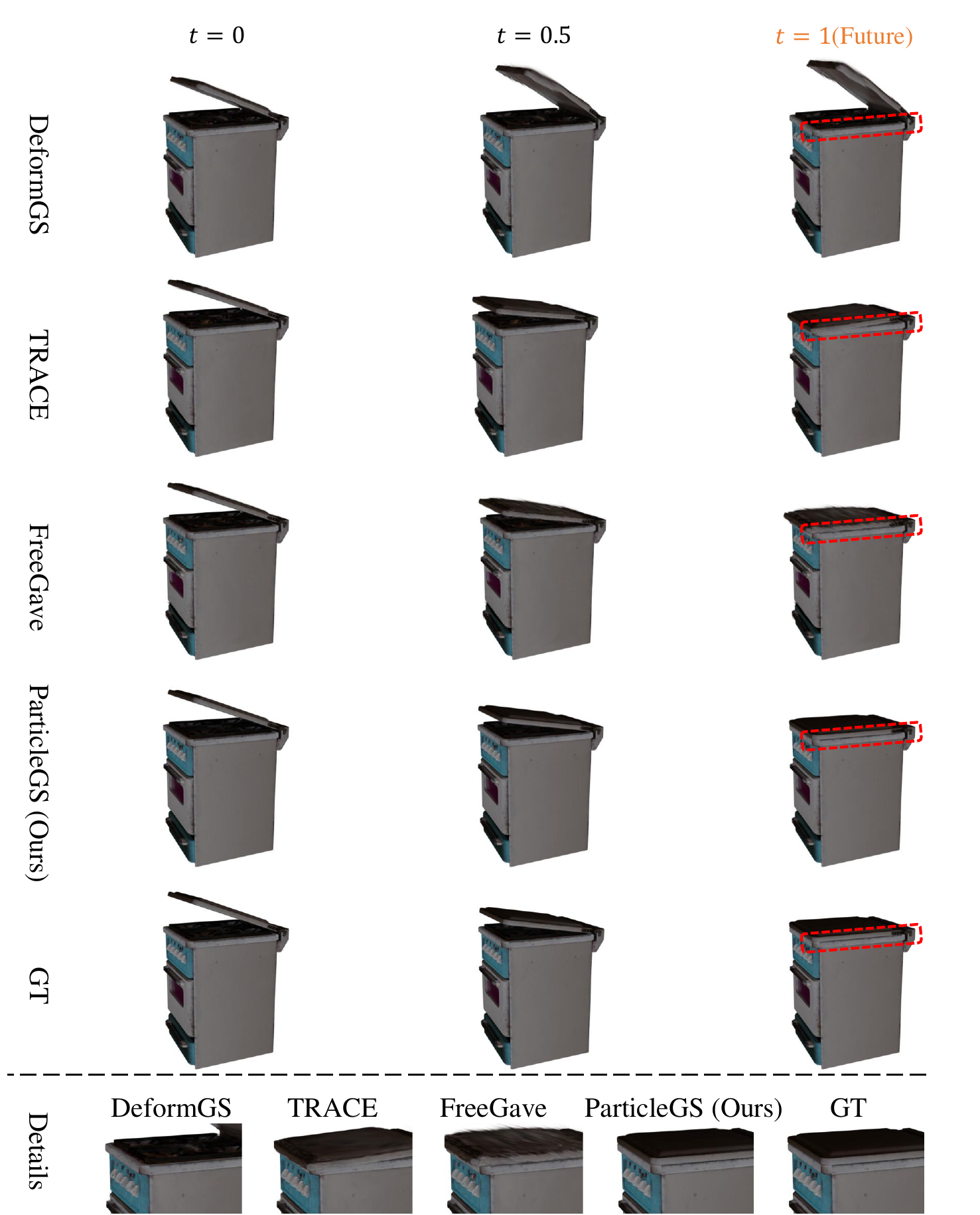}
    \caption{Additional qualitative comparisons on the Dynamic Multipart Dataset.}
    \label{app_fig5}
\end{figure*}

\begin{figure*}
    \centering
    \includegraphics[width=1.0\linewidth]{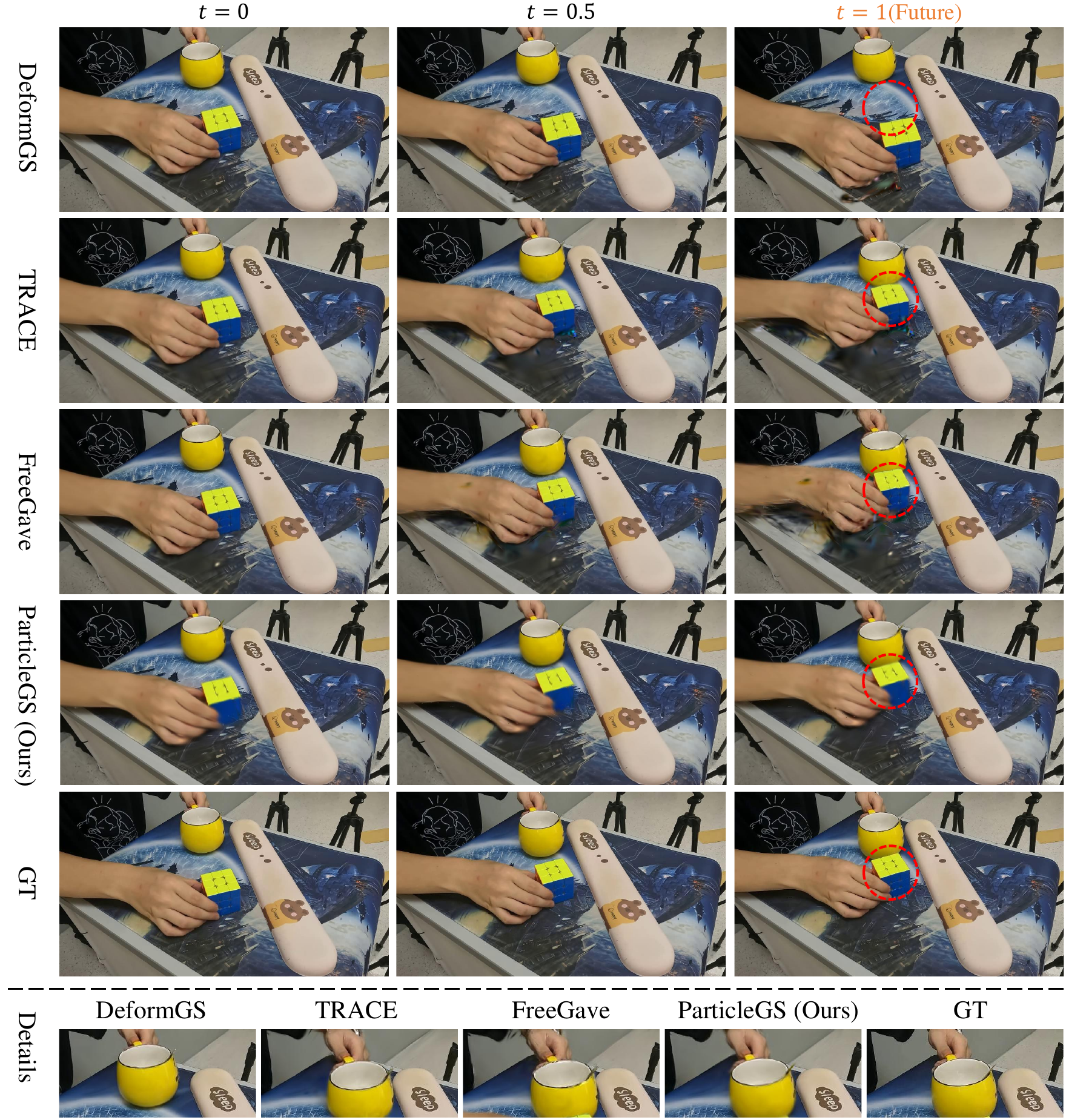}
    \caption{Additional qualitative comparisons on the FreeGave-GoPro Dataset}
    \label{app_fig7}
\end{figure*}

We provide additional visual comparisons across various datasets in Fig.~\ref{app_fig3} through Fig.~\ref{app_fig7}. The \textbf{red bounding boxes} in the figures mark the regions with ground truth motion, serving as a visual reference for the object's actual position and movement trajectory. The \textbf{bottom section} of each sub-figure displays details of specific regions. 

\section{Full Quantitative Results}
\label{sec:quantitative_results}

We report the complete quantitative results for both reconstruction and extrapolation tasks. Detailed metrics, including PSNR, SSIM, and LPIPS, are provided for every scene in the datasets to ensure a thorough evaluation.

\subsection{Full Results on Dynamic Object Dataset}
Table~\ref{tab:full_dynamic_obj} presents the detailed quantitative comparisons on the Dynamic Object Dataset, which includes scenes such as Falling Ball, Bat, Fan, Telescope, Shark, and Whale. 

\subsection{Full Results on Dynamic Indoor Scene Dataset}
The results for the Dynamic Indoor Scene Dataset are summarized in Table~\ref{tab:full_indoor}. This dataset features complex background geometries and lighting conditions, represented by scenes like Gnome House, Chessboard, Factory, and Dining Table. 

\subsection{Full Results on Dynamic Multipart Dataset} We also evaluate our method on the Dynamic Multipart Dataset, which includes objects exhibiting distinct part-level motion patterns, such as Folding Chair, Satellite, Stove, and Hyperbolic. The full quantitative results are shown in Table~\ref{tab:full_misc}.

\subsection{Full Results on FreeGave-GoPro Dataset}
Finally, Table~\ref{tab:full_freegave} details the performance on the FreeGave-GoPro Dataset, covering real-world captured scenes such as Box, Collision, Wrist Rest, Hammer, and Pen. 

\begin{table*}
  \caption{Full quantitative results on the Dynamic Objects, rendered at the source resolution. \textbf{Bold} and \underline{underline} indicate the best and second best performance.}
  \centering
  \resizebox{\linewidth}{!}{
  \begin{tabular}{lcccccccccccc}
    \toprule
    \multirow{3}{*}{Method} 
      & \multicolumn{6}{c}{Falling Ball} 
      & \multicolumn{6}{c}{Bat} \\
    \cmidrule(lr){2-7} \cmidrule(lr){8-13}
      & \multicolumn{3}{c}{Reconstruction} & \multicolumn{3}{c}{Extrapolation} 
      & \multicolumn{3}{c}{Reconstruction} & \multicolumn{3}{c}{Extrapolation} \\
    \cmidrule(r){2-4} \cmidrule(r){5-7} \cmidrule(r){8-10} \cmidrule(r){11-13}
      & PSNR$\uparrow$ & SSIM$\uparrow$ & LPIPS$\downarrow$ & PSNR$\uparrow$ & SSIM$\uparrow$ & LPIPS$\downarrow$  
      & PSNR$\uparrow$ & SSIM$\uparrow$ & LPIPS$\downarrow$ & PSNR$\uparrow$ & SSIM$\uparrow$ & LPIPS$\downarrow$ \\
    \midrule
    HexPlane & 19.754  & 0.816  & 0.215  & 20.166  & 0.906  & 0.130  & 22.099  & 0.953  & 0.080  & 22.422  & 0.954  & 0.061  \\
    TiNeuVox & 24.323  & 0.894  & 0.167  & 20.474  & 0.940  & 0.098  & 17.611  & 0.946  & 0.076  & 17.139  & 0.947  & 0.070  \\
    DeformGS & 17.736  & 0.830  & 0.156  & 26.384  & 0.945  & 0.060  & \underline{47.034}  & \underline{0.996}  & 0.004  & 27.109  & 0.944  & 0.034  \\
    Grid4D & 19.119  & 0.821  & 0.153  & 24.926  & 0.925  & 0.069  & \textbf{47.443}  & \textbf{0.997}  & 0.004  & 29.551  & 0.953  & 0.290  \\
    NVFi  & 34.900  & 0.969  & 0.068  & 27.013  & 0.966  & 0.074  & 24.648  & 0.970  & 0.044  & 25.529  & 0.973  & 0.040  \\
    GSPrediction & 24.549  & 0.910  & 0.154  & 16.790  & 0.830  & 0.238  & 43.290  & \textbf{0.997}  & \textbf{0.002}  & 23.495  & 0.952  & 0.075  \\
    TRACE & \underline{41.077}  & \underline{0.995}  & 0.017  & \textbf{38.126}  & \underline{0.993}  & 0.020  & 39.269  & 0.995  & 0.006  & 28.890  & \underline{0.982}  & 0.016  \\
    FreeGave & \textbf{42.248}  & \textbf{0.996}  & \underline{0.015}  & 37.460  & \textbf{0.994}  & \underline{0.015}  & 39.516  & 0.995  & 0.006  & \underline{30.907}  & 0.980  & \underline{0.014}  \\
    \midrule
    \textbf{ParticleGS} & 38.275  & 0.991  & \textbf{0.010}  & \underline{38.067}  & 0.990  & \textbf{0.013}  & 43.232  & \textbf{0.997}  & \underline{0.003}  & \textbf{36.348}  & \textbf{0.984}  & \textbf{0.011}  \\
    \bottomrule
  \end{tabular}
  }

    \vspace{0.4cm}
    
  \centering
  \resizebox{\linewidth}{!}{
  \begin{tabular}{lcccccccccccc}
    \toprule
    \multirow{3}{*}{Method} 
      & \multicolumn{6}{c}{Fan} 
      & \multicolumn{6}{c}{Telescope} \\
    \cmidrule(lr){2-7} \cmidrule(lr){8-13}
      & \multicolumn{3}{c}{Reconstruction} & \multicolumn{3}{c}{Extrapolation} 
      & \multicolumn{3}{c}{Reconstruction} & \multicolumn{3}{c}{Extrapolation} \\
    \cmidrule(r){2-4} \cmidrule(r){5-7} \cmidrule(r){8-10} \cmidrule(r){11-13}
      & PSNR$\uparrow$ & SSIM$\uparrow$ & LPIPS$\downarrow$ & PSNR$\uparrow$ & SSIM$\uparrow$ & LPIPS$\downarrow$  
      & PSNR$\uparrow$ & SSIM$\uparrow$ & LPIPS$\downarrow$ & PSNR$\uparrow$ & SSIM$\uparrow$ & LPIPS$\downarrow$ \\
    \midrule
    HexPlane & 18.424  & 0.842  & 0.190  & 19.998  & 0.887  & 0.106  & 21.814  & 0.917  & 0.149  & 23.667  & 0.931  & 0.082  \\
    TiNeuVox & 15.775  & 0.855  & 0.189  & 21.307  & 0.919  & 0.081  & 23.088  & 0.949  & 0.084  & 20.801  & 0.925  & 0.071  \\
    DeformGS & \underline{38.083}  & 0.973  & 0.026  & 24.473  & 0.930  & 0.047  & 38.435  & 0.992  & \underline{0.005}  & 23.186  & 0.934  & 0.037  \\
    Grid4D & 37.031  & 0.967  & 0.030  & 26.720  & 0.941  & 0.044  & 38.488  & 0.992  & 0.006  & 25.696  & 0.942  & 0.033  \\
    NVFi  & 27.748  & 0.953  & 0.059  & 27.347  & 0.957  & 0.054  & 26.889  & 0.955  & 0.051  & 26.861  & 0.955  & 0.053  \\
    GSPrediction & 34.864  & \underline{0.976}  & \textbf{0.020}  & 20.747  & 0.935  & 0.054  & 36.191  & 0.992  & \textbf{0.004}  & 22.018  & 0.939  & 0.043  \\
    TRACE & 34.038  & 0.975  & 0.023  & \textbf{36.779}  & \textbf{0.984}  & \textbf{0.014}  & \textbf{38.839}  & \textbf{0.994}  & 0.006  & 35.725  & \underline{0.980}  & 0.006  \\
    FreeGave & 34.158  & 0.975  & 0.022  & 33.987  & \underline{0.974}  & \underline{0.016}  & 38.737  & \textbf{0.994}  & 0.006  & \underline{36.407}  & \textbf{0.991}  & \underline{0.005}  \\
    \midrule
    \textbf{ParticleGS} & \textbf{38.852}  & \textbf{0.979}  & \underline{0.021}  & \underline{34.235}  & \underline{0.974}  & 0.023  & \underline{38.770}  & \underline{0.993}  & \textbf{0.004}  & \textbf{38.294}  & \textbf{0.991}  & \textbf{0.004}  \\
    \bottomrule
  \end{tabular}
  }

    \vspace{0.4cm}

    \centering
  \resizebox{\linewidth}{!}{
  \begin{tabular}{lcccccccccccc}
    \toprule
    \multirow{3}{*}{Method} 
      & \multicolumn{6}{c}{Shark} 
      & \multicolumn{6}{c}{Whale} \\
    \cmidrule(lr){2-7} \cmidrule(lr){8-13}
      & \multicolumn{3}{c}{Reconstruction} & \multicolumn{3}{c}{Extrapolation} 
      & \multicolumn{3}{c}{Reconstruction} & \multicolumn{3}{c}{Extrapolation} \\
    \cmidrule(r){2-4} \cmidrule(r){5-7} \cmidrule(r){8-10} \cmidrule(r){11-13}
      & PSNR$\uparrow$ & SSIM$\uparrow$ & LPIPS$\downarrow$ & PSNR$\uparrow$ & SSIM$\uparrow$ & LPIPS$\downarrow$  
      & PSNR$\uparrow$ & SSIM$\uparrow$ & LPIPS$\downarrow$ & PSNR$\uparrow$ & SSIM$\uparrow$ & LPIPS$\downarrow$ \\
    \midrule
    HexPlane & 22.736 & 0.956 & 0.064 & 24.180 & 0.962 & 0.041 & 24.684 & 0.976 & 0.049 & 24.075 & 0.964 & 0.049 \\
    TiNeuVox & 19.592 & 0.951 & 0.058 & 16.683 & 0.933 & 0.096 & 21.281 & 0.969 & 0.052 & 20.671 & 0.945 & 0.063 \\
    DeformGS & \underline{40.492} & \textbf{0.995} & \underline{0.008} & 29.159 & 0.966 & 0.022 & \underline{42.475} & \textbf{0.996} & \textbf{0.004} & 26.666 & 0.959 & 0.027 \\
    Grid4D & \textbf{40.752} & \textbf{0.995} & \textbf{0.007} & 30.904 & 0.966 & 0.021 & \textbf{42.666} & \textbf{0.996} & \textbf{0.004} & 28.246 & 0.967 & 0.020 \\
    NVFi & 31.374 & \underline{0.982} & 0.035 & 28.649 & 0.979 & 0.033 & 30.778 & 0.984 & 0.025 & 25.529 & 0.977 & 0.029 \\
    GSPrediction & 40.015 & \textbf{0.995} & 0.010 & 30.280 & 0.971 & 0.032 & 39.540 & \textbf{0.996} & \underline{0.005} & 25.596 & 0.965 & 0.046 \\
    TRACE & 38.637 & \textbf{0.995} & 0.009 & 30.547 & \underline{0.982} & 0.012 & 36.177 & 0.994 & 0.006 & 30.068 & \underline{0.980} & \underline{0.013} \\
    FreeGave & 39.378 & \textbf{0.995} & 0.009 & \underline{31.639} & 0.979 & \underline{0.011} & 37.788 & 0.994 & 0.006 & \underline{31.395} & \underline{0.980} & \textbf{0.012} \\
    \midrule
    \textbf{ParticleGS} & 40.322 & \textbf{0.995} & \underline{0.008} & \textbf{37.475} & \textbf{0.983} & \textbf{0.010} & 39.241 & \underline{0.995} & \underline{0.005} & \textbf{34.236} & \textbf{0.983} & \textbf{0.012} \\
    \bottomrule
  \end{tabular}
  }
  \label{tab:full_dynamic_obj}
\end{table*}

\begin{table*}
  \caption{Full quantitative results on Dynamic Indoor Scene dataset, rendered at the source resolution. \textbf{Bold} and \underline{underline} indicate the best and second best performance.}
  \centering
  \resizebox{\linewidth}{!}{
  \begin{tabular}{lcccccccccccc}
    \toprule
    \multirow{3}{*}{Method} 
      & \multicolumn{6}{c}{Gnome House} 
      & \multicolumn{6}{c}{Chessboard} \\
    \cmidrule(lr){2-7} \cmidrule(lr){8-13}
      & \multicolumn{3}{c}{Reconstruction} & \multicolumn{3}{c}{Extrapolation} 
      & \multicolumn{3}{c}{Reconstruction} & \multicolumn{3}{c}{Extrapolation} \\
    \cmidrule(r){2-4} \cmidrule(r){5-7} \cmidrule(r){8-10} \cmidrule(r){11-13}
      & PSNR$\uparrow$ & SSIM$\uparrow$ & LPIPS$\downarrow$ & PSNR$\uparrow$ & SSIM$\uparrow$ & LPIPS$\downarrow$  
      & PSNR$\uparrow$ & SSIM$\uparrow$ & LPIPS$\downarrow$ & PSNR$\uparrow$ & SSIM$\uparrow$ & LPIPS$\downarrow$ \\
    \midrule
    HexPlane & 18.060 & 0.419 & 0.650 & 23.341 & 0.601 & 0.530 & 18.054 & 0.515 & 0.617 & 22.034 & 0.680 & 0.500 \\
    TiNeuVox & \textbf{24.644} & \textbf{0.663} & 0.385 & 21.359 & 0.665 & 0.357 & \underline{23.847} & \underline{0.698} & 0.392 & 20.088 & 0.676 & 0.408 \\
    DeformGS & 21.855 & 0.611 & \underline{0.341} & 22.728 & 0.731 & \underline{0.196} & 20.688 & 0.659 & 0.443 & 20.484 & 0.773 & 0.280 \\
    Grid4D & 22.418 & 0.588 & 0.360 & 21.185 & 0.696 & 0.258 & 21.545 & 0.667 & 0.437 & 19.713 & 0.741 & 0.331 \\
    NVFi & 16.857 & 0.435 & 0.687 & 24.087 & 0.615 & 0.546 & 17.515 & 0.551 & 0.651 & 22.571 & 0.672 & 0.539 \\
    GSPrediction & 16.113 & 0.556 & 0.632 & 20.098 & 0.641 & 0.371 & 17.665 & 0.651 & 0.470 & 20.033 & 0.724 & 0.317 \\
    TRACE & 22.968 & 0.642 & 0.351 & \underline{29.325} & \underline{0.840} & 0.209 & 22.969 & 0.642 & \underline{0.352} & \underline{29.326} & 0.840 & \underline{0.209} \\
    FreeGave & 19.178 & 0.543 & 0.425 & 28.962 & 0.827 & 0.217 & 19.905 & 0.674 & 0.471 & 26.339 & \underline{0.862} & 0.251 \\
    \midrule
    \textbf{ParticleGS} & \underline{24.481} & \underline{0.662} & \textbf{0.307} & \textbf{31.167} & \textbf{0.843} & \textbf{0.160} & \textbf{25.449} & \textbf{0.785} & \textbf{0.287} & \textbf{29.529} & \textbf{0.898} & \textbf{0.153} \\
    \bottomrule
  \end{tabular}
  }

    \vspace{0.4cm}
    
  \centering
  \resizebox{\linewidth}{!}{
  \begin{tabular}{lcccccccccccc}
    \toprule
    \multirow{3}{*}{Method} 
      & \multicolumn{6}{c}{Factory} 
      & \multicolumn{6}{c}{Dining Table} \\
    \cmidrule(lr){2-7} \cmidrule(lr){8-13}
      & \multicolumn{3}{c}{Reconstruction} & \multicolumn{3}{c}{Extrapolation} 
      & \multicolumn{3}{c}{Reconstruction} & \multicolumn{3}{c}{Extrapolation} \\
    \cmidrule(r){2-4} \cmidrule(r){5-7} \cmidrule(r){8-10} \cmidrule(r){11-13}
      & PSNR$\uparrow$ & SSIM$\uparrow$ & LPIPS$\downarrow$ & PSNR$\uparrow$ & SSIM$\uparrow$ & LPIPS$\downarrow$  
      & PSNR$\uparrow$ & SSIM$\uparrow$ & LPIPS$\downarrow$ & PSNR$\uparrow$ & SSIM$\uparrow$ & LPIPS$\downarrow$ \\
    \midrule
    HexPlane & 19.109 & 0.479 & 0.637 & 24.266 & 0.677 & 0.470 & 16.687 & 0.516 & 0.601 & 23.127 & 0.725 & 0.395 \\
    TiNeuVox & \underline{24.857} & \underline{0.705} & \underline{0.346} & 22.467 & 0.747 & 0.292 & \underline{21.476} & \underline{0.658} & \textbf{0.368} & 20.529 & 0.768 & 0.276 \\
    DeformGS & 21.703 & 0.621 & 0.376 & 23.471 & 0.787 & \underline{0.196} & 15.832 & 0.488 & 0.621 & 21.244 & 0.784 & 0.258 \\
    Grid4D & 21.822 & 0.631 & 0.377 & 22.119 & 0.765 & 0.235 & 16.407 & 0.475 & 0.595 & 19.516 & 0.741 & 0.313 \\
    NVFi & 18.889 & 0.503 & 0.655 & 25.415 & 0.690 & 0.496 & 15.206 & 0.485 & 0.681 & 22.790 & 0.711 & 0.428 \\
    GSPrediction & 17.388 & 0.588 & 0.609 & 21.723 & 0.713 & 0.308 & 13.451 & 0.546 & 0.688 & 18.214 & 0.679 & 0.374 \\
    TRACE & 22.012 & 0.581 & 0.426 & 27.177 & 0.817 & 0.234 & \textbf{23.434} & \textbf{0.705} & \underline{0.380} & \underline{32.078} & \textbf{0.900} & 0.169 \\
    FreeGave & 19.448 & 0.498 & 0.485 & \underline{28.190} & \underline{0.827} & 0.221 & 20.194 & 0.643 & 0.433 & \textbf{32.443} & \textbf{0.900} & \underline{0.159} \\
    \midrule
    \textbf{ParticleGS} & \textbf{31.627} & \textbf{0.912} & \textbf{0.126} & \textbf{32.679} & \textbf{0.943} & \textbf{0.077} & 20.461 & 0.636 & 0.444 & 31.037 & \underline{0.886} & \textbf{0.143} \\
    \bottomrule
  \end{tabular}
  }
  \label{tab:full_indoor}
\end{table*}

\begin{table*}
  \caption{Full quantitative results on the Dynamic Multipart dataset, rendered at the source resolution. \textbf{Bold} and \underline{underline} indicate the best and second best performance.}
  \centering
  \resizebox{\linewidth}{!}{
  \begin{tabular}{lcccccccccccc}
    \toprule
    \multirow{3}{*}{Method} 
      & \multicolumn{6}{c}{Folding Chair} 
      & \multicolumn{6}{c}{Satellite} \\
    \cmidrule(lr){2-7} \cmidrule(lr){8-13}
      & \multicolumn{3}{c}{Reconstruction} & \multicolumn{3}{c}{Extrapolation} 
      & \multicolumn{3}{c}{Reconstruction} & \multicolumn{3}{c}{Extrapolation} \\
    \cmidrule(r){2-4} \cmidrule(r){5-7} \cmidrule(r){8-10} \cmidrule(r){11-13}
      & PSNR$\uparrow$ & SSIM$\uparrow$ & LPIPS$\downarrow$ & PSNR$\uparrow$ & SSIM$\uparrow$ & LPIPS$\downarrow$  
      & PSNR$\uparrow$ & SSIM$\uparrow$ & LPIPS$\downarrow$ & PSNR$\uparrow$ & SSIM$\uparrow$ & LPIPS$\downarrow$ \\
    \midrule
    DeformGS & \underline{41.562} & \textbf{0.994} & \textbf{0.009} & 18.194 & 0.847 & 0.123 & 35.884 & 0.988 & \textbf{0.010} & 29.454 & 0.980 & 0.014 \\
    FreeGave & 39.086 & \textbf{0.994} & \underline{0.010} & 26.043 & 0.951 & 0.037 & \underline{37.155} & \textbf{0.991} & \textbf{0.010} & \underline{34.476} & \textbf{0.989} & \textbf{0.007} \\
    TRACE & 38.286 & \underline{0.993} & 0.011 & \underline{28.724} & \underline{0.974} & \underline{0.023} & 36.647 & \underline{0.990} & \textbf{0.010} & 33.612 & \underline{0.983} & \underline{0.009} \\
    \midrule
    \textbf{ParticleGS} & \textbf{41.620} & \textbf{0.994} & \textbf{0.009} & \textbf{31.512} & \textbf{0.977} & \textbf{0.015} & \textbf{38.697} & \textbf{0.991} & \textbf{0.010} & \textbf{34.562} & 0.981 & 0.017 \\
    \bottomrule
  \end{tabular}
  }

    \vspace{0.4cm}
    
  \centering
  \resizebox{\linewidth}{!}{
  \begin{tabular}{lcccccccccccc}
    \toprule
    \multirow{3}{*}{Method} 
      & \multicolumn{6}{c}{Stove} 
      & \multicolumn{6}{c}{Hyperbolic} \\
    \cmidrule(lr){2-7} \cmidrule(lr){8-13}
      & \multicolumn{3}{c}{Reconstruction} & \multicolumn{3}{c}{Extrapolation} 
      & \multicolumn{3}{c}{Reconstruction} & \multicolumn{3}{c}{Extrapolation} \\
    \cmidrule(r){2-4} \cmidrule(r){5-7} \cmidrule(r){8-10} \cmidrule(r){11-13}
      & PSNR$\uparrow$ & SSIM$\uparrow$ & LPIPS$\downarrow$ & PSNR$\uparrow$ & SSIM$\uparrow$ & LPIPS$\downarrow$  
      & PSNR$\uparrow$ & SSIM$\uparrow$ & LPIPS$\downarrow$ & PSNR$\uparrow$ & SSIM$\uparrow$ & LPIPS$\downarrow$ \\
    \midrule
    DeformGS & \textbf{40.169} & 0.988 & \underline{0.027} & 30.694 & 0.953 & 0.030 & \textbf{36.105} & 0.983 & \textbf{0.031} & 33.614 & 0.987 & 0.015 \\
    FreeGave & 35.578 & \underline{0.989} & 0.028 & 32.291 & \underline{0.989} & 0.021 & 32.747 & \textbf{0.985} & 0.034 & \textbf{41.294} & \textbf{0.995} & \textbf{0.011} \\
    TRACE & 36.201 & \textbf{0.990} & 0.028 & \underline{33.869} & \underline{0.989} & \underline{0.016} & 33.942 & \textbf{0.985} & 0.036 & 37.628 & \underline{0.990} & \underline{0.012} \\
    \midrule
    \textbf{ParticleGS} & \underline{39.115} & \underline{0.989} & \textbf{0.024} & \textbf{40.069} & \textbf{0.991} & \textbf{0.013} & \underline{34.998} & \underline{0.984} & \underline{0.032} & \underline{38.399} & \underline{0.990} & \underline{0.012} \\
    \bottomrule
  \end{tabular}
  }
  \label{tab:full_misc}
\end{table*}

\begin{table*}
  \caption{Full quantitative results on the FreeGave-GoPro dataset, rendered at the source resolution. \textbf{Bold} and \underline{underline} indicate the best and second best performance.}
  \centering
  \resizebox{\linewidth}{!}{
  \begin{tabular}{lcccccccccccc}
    \toprule
    \multirow{3}{*}{Method} 
      & \multicolumn{6}{c}{Box} 
      & \multicolumn{6}{c}{Collision} \\
    \cmidrule(lr){2-7} \cmidrule(lr){8-13}
      & \multicolumn{3}{c}{Reconstruction} & \multicolumn{3}{c}{Extrapolation} 
      & \multicolumn{3}{c}{Reconstruction} & \multicolumn{3}{c}{Extrapolation} \\
    \cmidrule(r){2-4} \cmidrule(r){5-7} \cmidrule(r){8-10} \cmidrule(r){11-13}
      & PSNR$\uparrow$ & SSIM$\uparrow$ & LPIPS$\downarrow$ & PSNR$\uparrow$ & SSIM$\uparrow$ & LPIPS$\downarrow$  
      & PSNR$\uparrow$ & SSIM$\uparrow$ & LPIPS$\downarrow$ & PSNR$\uparrow$ & SSIM$\uparrow$ & LPIPS$\downarrow$ \\
    \midrule
    DeformGS & 19.505 & 0.810 & 0.221 & 25.754 & 0.900 & 0.139 & \underline{20.012} & \underline{0.840} & \textbf{0.209} & 18.986 & 0.810 & 0.251 \\
    TRACE & \textbf{20.203} & \textbf{0.833} & \textbf{0.195} & 27.987 & 0.912 & \underline{0.131} & \textbf{20.242} & \textbf{0.841} & \textbf{0.209} & 23.447 & 0.858 & 0.193 \\
    FreeGave & 19.800 & 0.830 & \underline{0.197} & \underline{28.661} & \underline{0.913} & \underline{0.131} & 19.608 & \underline{0.840} & 0.212 & \underline{24.188} & \underline{0.867} & \underline{0.187} \\
    \midrule
    \textbf{ParticleGS} & \underline{20.122} & \underline{0.831} & \textbf{0.195} & \textbf{29.057} & \textbf{0.919} & \textbf{0.102} & 19.759 & \textbf{0.841} & \underline{0.211} & \textbf{24.239} & \textbf{0.901} & \textbf{0.157} \\
    \bottomrule
  \end{tabular}
  }

    \vspace{0.4cm}
    
  \centering
  \resizebox{\linewidth}{!}{
  \begin{tabular}{lcccccccccccc}
    \toprule
    \multirow{3}{*}{Method} 
      & \multicolumn{6}{c}{Wrist Rest} 
      & \multicolumn{6}{c}{Hammer} \\
    \cmidrule(lr){2-7} \cmidrule(lr){8-13}
      & \multicolumn{3}{c}{Reconstruction} & \multicolumn{3}{c}{Extrapolation} 
      & \multicolumn{3}{c}{Reconstruction} & \multicolumn{3}{c}{Extrapolation} \\
    \cmidrule(r){2-4} \cmidrule(r){5-7} \cmidrule(r){8-10} \cmidrule(r){11-13}
      & PSNR$\uparrow$ & SSIM$\uparrow$ & LPIPS$\downarrow$ & PSNR$\uparrow$ & SSIM$\uparrow$ & LPIPS$\downarrow$  
      & PSNR$\uparrow$ & SSIM$\uparrow$ & LPIPS$\downarrow$ & PSNR$\uparrow$ & SSIM$\uparrow$ & LPIPS$\downarrow$ \\
    \midrule
    DeformGS & 19.498 & 0.829 & 0.222 & 19.007 & 0.808 & 0.244 & 20.067 & 0.835 & 0.198 & 23.992 & 0.896 & 0.152 \\
    TRACE & 19.079 & \underline{0.833} & \underline{0.209} & 22.488 & 0.845 & \underline{0.204} & 20.077 & \textbf{0.842} & 0.195 & \textbf{28.924} & \textbf{0.919} & \underline{0.137} \\
    FreeGave & \textbf{19.807} & \textbf{0.835} & \textbf{0.208} & \textbf{23.456} & \textbf{0.853} & \textbf{0.197} & \textbf{20.407} & \underline{0.840} & \underline{0.194} & 28.521 & \underline{0.917} & \underline{0.137} \\
    \midrule
    \textbf{ParticleGS} & \underline{19.613} & \textbf{0.835} & \textbf{0.208} & \underline{23.000} & \underline{0.846} & 0.208 & \underline{20.301} & \textbf{0.842} & \textbf{0.193} & \underline{28.816} & \underline{0.917} & \textbf{0.136} \\
    \bottomrule
  \end{tabular}
  }

    \vspace{0.4cm}

    \centering
  \resizebox{\linewidth}{!}{
  \begin{tabular}{lcccccccccccc}
    \toprule
    \multirow{3}{*}{Method} 
      & \multicolumn{6}{c}{Pen1} 
      & \multicolumn{6}{c}{Pen2} \\
    \cmidrule(lr){2-7} \cmidrule(lr){8-13}
      & \multicolumn{3}{c}{Reconstruction} & \multicolumn{3}{c}{Extrapolation} 
      & \multicolumn{3}{c}{Reconstruction} & \multicolumn{3}{c}{Extrapolation} \\
    \cmidrule(r){2-4} \cmidrule(r){5-7} \cmidrule(r){8-10} \cmidrule(r){11-13}
      & PSNR$\uparrow$ & SSIM$\uparrow$ & LPIPS$\downarrow$ & PSNR$\uparrow$ & SSIM$\uparrow$ & LPIPS$\downarrow$  
      & PSNR$\uparrow$ & SSIM$\uparrow$ & LPIPS$\downarrow$ & PSNR$\uparrow$ & SSIM$\uparrow$ & LPIPS$\downarrow$ \\
    \midrule
    DeformGS & \textbf{21.180} & 0.851 & \textbf{0.205} & 20.577 & 0.865 & 0.197 & \textbf{20.372} & 0.824 & 0.217 & 21.674 & 0.870 & 0.190 \\
    TRACE & 20.977 & \underline{0.854} & \textbf{0.205} & 26.703 & \underline{0.917} & 0.149 & \underline{20.040} & \underline{0.829} & \textbf{0.214} & 25.939 & 0.899 & 0.167 \\
    FreeGave & 20.130 & 0.853 & \underline{0.206} & \underline{27.673} & \textbf{0.923} & \underline{0.143} & 19.968 & \textbf{0.830} & \underline{0.215} & \underline{26.563} & \underline{0.903} & \underline{0.162} \\
    \midrule
    \textbf{ParticleGS} & \underline{21.079} & \textbf{0.856} & \textbf{0.205} & \textbf{27.746} & \textbf{0.923} & \textbf{0.140} & 19.824 & \underline{0.829} & \underline{0.215} & \textbf{27.879} & \textbf{0.910} & \textbf{0.158} \\
    \bottomrule
  \end{tabular}
  }
  \label{tab:full_freegave}
\end{table*}

\end{document}